\documentclass{article}

\usepackage[preprint]{neurips_2024}

\usepackage[utf8]{inputenc} 
\usepackage[T1]{fontenc}    
\usepackage{hyperref}       
\usepackage{url}            
\usepackage{booktabs}       
\usepackage{amsfonts}       
\usepackage{nicefrac}       
\usepackage{microtype}      
\usepackage{xcolor}         

\usepackage{graphicx}

\usepackage{outlines}

\usepackage{amsmath}
\usepackage{amssymb}
\usepackage{mathtools}
\usepackage{amsthm}

\usepackage{algorithm}
\usepackage{algorithmic}

\title{Biologically-Motivated Learning Model for Instructed Visual Processing}

\author{%
  Roy Abel \\
  Weizmann Institute of Science \\
  \texttt{roy.abel@weizmann.ac.il} \\
  \And
  Shimon Ullman \\
  Weizmann Institute of Science \\
  \texttt{shimon.ullman@weizmann.ac.il} \\
}

\begin{document}

\maketitle

\begin{abstract}

As part of understanding how the brain learns, ongoing work seeks to combine biological knowledge and current artificial intelligence (AI) modeling in an attempt to find an efficient biologically plausible learning scheme. Current models of biologically plausible learning often use a cortical-like combination of bottom-up (BU) and top-down (TD) processing, where the TD part carries feedback signals used for learning. However, in the visual cortex, the TD pathway plays a second major role of visual attention, by guiding the visual process to locations and tasks of interest. A biological model should therefore combine the two tasks, and learn to guide the visual process. We introduce a model that uses a cortical-like combination of BU and TD processing that naturally integrates the two major functions of the TD stream. The integrated model is obtained by an appropriate connectivity pattern between the BU and TD streams, a novel processing cycle that uses the TD part twice, and the use of 'Counter-Hebb' learning that operates across the streams. We show that the 'Counter-Hebb' mechanism can provide an exact backpropagation synaptic modification. We further demonstrate the model's ability to guide the visual stream to perform a task of interest, achieving competitive performance compared with AI models on standard multi-task learning benchmarks. The successful combination of learning and visual guidance could provide a new view on combining BU and TD processing in human vision, and suggests possible directions for both biologically plausible models and artificial instructed models, such as vision-language models (VLMs).

\end{abstract}

\section{Introduction}


Understanding how the human brain learns has been a longstanding pursuit in Neuroscience and Artificial Intelligence (AI). An extensive research area at the intersection of AI and Neuroscience has been the development of biologically plausible models of cortical learning, in particular in visual processing \cite{ernoult2022towards, bozkurt2024correlative}. A Detailed biological model is an ultimate goal, and at present the main focus has been on possible schemes for modifying synaptic weights during learning, e.g. how the modifications are determined, and how they propagate throughout the cortical network (the credit assignment problem) \cite{whittington2019theories}. 

Models of biologically plausible learning often use a combination of feedforward (bottom-up) and feedback (top-down) pathways, where the top-down (TD) stream plays a major role in the learning process, similar to the cortical structure in the human brain \cite{lillicrap2020backpropagation, song2021convergence}. However, a notable difference exists between these models and the cortex: while biological models primarily use the TD stream for propagating feedback signals that are used in synaptic modifications, the TD stream in the cortex has an additional major role in perception by directing TD attention \cite{manita2015top}. Studies indicate that the TD stream guides the visual processes to selected locations and tasks, thereby actively influencing the neural activity of the bottom-up (BU) stream \cite{gilbert2013top}, creating task-dependent representations in the cortex \cite{harel2014task}. Therefore, current biological plausible models are being criticized for not involving the TD stream in ongoing visual processes \cite{lillicrap2020backpropagation}, and it is still an open research question how to incorporate the TD stream to participate in the feedforward BU processing \cite{zagha2020shaping, kreiman2020beyond}. 

The current work addresses this deficiency and suggests, for the first time to our knowledge, a biologically plausible learning model where the TD stream not only carries feedback signals, but also performs visual guidance. The model addresses two main challenges that must be addressed together: firstly, \textit{guidance}, understanding how the TD stream guides BU neural processing, and secondly, \textit{learning}, which requires solving the credit assignment problem in this setting of guided processing. 

In this work, we propose a model where the TD component accounts for both directing the BU visual process and for determining synaptic modifications. As a result, the model may offer a more accurate description of the bi-directional cortical processing and learning, compared with existing models. For learning, we suggest the use of a 'Counter-Hebb' mechanism, which is a modification of the classical Hebbian learning \cite{hebb2005organization}, and we show that the model can provide an exact version of backpropagation (BP) synaptic modification \cite{rumelhart1986learning}. Regarding its biological plausibility, we address the weight symmetry problem between forward and backward paths, as well as using local synaptic updates that depend only on neurons associated with the modified synapse. Furthermore, our method offers a possible solution to the long-standing challenge of integrating the TD stream into the ongoing visual processing \cite{zagha2020shaping, kreiman2020beyond, lillicrap2020backpropagation}. In the context of guidance, we show that the TD stream can be used for both learning and for directing the BU stream to perform tasks of interest, by selecting a sparse task-specific sub-network within the full BU network; The TD stream selects the appropriate sub-network based on the task, directing the BU processing to operate on this sub-network. We further show that this model presents competitive results on standard multi-task learning benchmarks.  

In addition to the brain-related aspects discussed above, it is noteworthy that the integration of guidance is becoming a central aspect of recent AI models, particularly in Large Language Models (LLMs) and Vision-Language Models (VLMs). A fundamental aspect of VLMs is that, similar to the brain, they include the capacity to direct the visual process to focus on tasks of interest, and they use the concept of instruction tuning \cite{huang2023visual, liu2024visual}. This guidance is obtained in VLMs by using an instruction that propagates through the language stream of the model, which then interacts with the visual part of the model. This parallel development in AI and brain modeling may prove beneficial for both our understanding of the human brain, as well as leading to the development of more advanced and human-like AI systems.

The key contributions of our work are: 
\begin{outline}
    \1 We propose the first biologically motivated learning model for instructed visual models.
    \1 We present a unified feedback mechanism that combines error propagation for synaptic learning and Top-Down attention to guide visual processing based on instructions, effectively modulating the feedforward neural activity in visual processing.
    \1 We suggest a Counter-Hebb learning procedure as a possible local synaptic modification that can perform the exact backpropagation learning.
\end{outline}

The code for reproducing the below experiments and creating BU-TD models for guided visual processing is available at \url{https://github.com/royabel/Top-Down-Networks}.

\section{Related work}

The fields of brain modeling and AI have beneficial interactions going in both directions \cite{yamins2016using, bowers2017parallel, yildirim2019integrative}. Particularly, the study of biologically plausible learning models aims to deepen our understanding of the learning mechanisms in the human brain and enhance learning techniques for artificial neural network models. While artificial models primarily employ the backpropagation (BP) algorithm for learning \cite{rumelhart1986learning}, a direct implementation of BP in biological models is generally considered biologically implausible \cite{whittington2019theories, lillicrap2020backpropagation}. Nevertheless, the integration of BP with biological principles, such as Hebb's plasticity rule \cite{hebb2005organization}, has inspired the development of diverse biologically plausible learning approaches. These methods are often compared to BP, aiming to achieve similar performance in a more biologically plausible way. For instance, Equilibrium Propagation methods \cite{scellier2017equilibrium} have been shown to produce weight updates equivalent to BP under specific conditions \cite{ernoult2019updates}, and approximate BP under others \cite{millidge2020activation}. Predictive Coding methods have been explored in supervised learning \cite{whittington2017approximation}, and have demonstrated the ability to both approximate the BP update \cite{millidge2022predictive} and perform the same weights update as BP \cite{song2020can, salvatori2022reverse}. However, the modifications necessary for these methods to approximate or be equivalent to BP reduce their biological plausibility \cite{rosenbaum2022relationship, golkar2022constrained}. 

Among biologically plausible approaches, the Feedback Alignment (FA) and Target Propagation (TP) approaches are most similar to our method. Like backpropagation, these approaches involve a forward stream (BU) that generates predictions based on an input signal, followed by a backward stream (TD) that propagates feedback information. While BP propagates gradients backward using the same weights as the forward path, FA methods propose propagating gradient-like signals through the TD stream via a separated set of weights, thus removing the symmetric weight structure of BP \cite{lillicrap2016random, nokland2016direct, song2021convergence}. The TP methods suggest propagating backward targets for the forward path instead of gradients \cite{bengio2014auto, lee2015difference, meulemans2020theoretical}. Both FA and TP methods can approximate the BP update under specific conditions \cite{akrout2019deep, ahmad2020gait, ernoult2022towards}. Nevertheless, current models lack the extensive BU-TD interactions observed in the brain, which are essential for guiding attention in visual processes \cite{harel2014task, manita2015top, wen2019goal, lillicrap2020backpropagation}.

\subsection{Guided visual processing}

Human cortical processing uses a combination of bottom-up (BU) and top-down (TD) processing streams. In the visual brain, the BU stream proceeds from low-level sensory regions to high-level, more cognitive areas, while in the TD stream processing flows in the opposite direction \cite{dehaene2021consciousness}. In human vision, the TD stream is involved in TD attention, guiding the visual process and directing it toward tasks of interest \cite{goddard2022spatial, shahdloo2022task}. For example, at the physiological level, it has been shown, in behaving primate studies, that given the same image, but with different tasks, the activation along the BU stream changes, modulated by TD activation, to focus on the instructed task \cite{gilbert2013top}.

The ability to guide visual processing to extract specific aspects of the image is essential because a single image encompasses a wealth of information regarding objects, their parts and sub-parts, their properties, and inter-relations. Consequently, for complex images, it becomes difficult to extract and represent all the possibly meaningful information through a single visual representation \cite{huang2023visual}. Empirical studies in artificial models have shown advantages to guiding the model's attention to selected locations or selected tasks compared with non-guided pure BU models \cite{tsotsos2021computational, pang2021tdaf, ullman2023human}. Furthermore, as opposed to earlier computer vision models, which relied solely on visual inputs without guidance, recent Vision Language Models (VLMs) have integrated guidance mechanisms into their visual processing \cite{bai2023qwen, zhu2023minigpt, liu2024visual, dai2024instructblip}. The processing of visual information in VLMs integrates a language stream that interacts with a visual stream and guides it to perform selected tasks. As a result, these models have been shown to have high generalization and zero-shot capabilities. Consequently, given the importance of guidance mechanisms in both the human cortex and AI models, developing biologically plausible learning models for guided processing is essential for both neuroscience implications and potential advancements of artificial models. While our focus in this paper centers on guided visual processing, it is worth noting that our method can be applied to guided processing in other domains as well. 

\section{The bottom-up top-down model}
\label{section: the BU-TD model}
In this section, we introduce the suggested structure of the Bottom-Up (BU) and Top-Down (TD) networks. A BU network with $L$ hidden layers is a function that maps an input vector $x \coloneqq h_0$ to an output vector $y$, such that for every layer $0 \leq l < L$: the hidden values are defined to be:
\begin{equation}
    h_{l+1} \coloneqq \sigma\left(f_{l+1}(h_0, h_1, ..., h_{l})\right)
\end{equation}
The functions $f_l$ are linear, and the activation function $\sigma$ is an element-wise function that may be non-linear. To predict an output, we use a prediction head $H_{pred}$, which is a small network, typically one to two layers, that maps the last hidden layer $h_L$ to the predicted output: $y = H_{pred}(h_L)$.

For a given BU network, we define a symmetric TD network (denoted with upper bars), to be the reversed architecture network that maps an input vector $\bar{y}$ to an output vector $\bar{x} \coloneqq \bar{h}_0$. The TD network is constructed based on the BU architecture as follows: The input (e.g. the prediction error) $\bar{y}$ is mapped to the top-level hidden layer $\bar{h}_L$ of the TD network via the TD prediction head: $\bar{h}_L = \bar{H}_{pred}(\bar{y})$, and then for every $0 \leq l < L$:
\begin{equation}
    \bar{h}_l \coloneqq \bar{\sigma}\left(\bar{f}_{l+1}(\bar{h}_{L}, \bar{h}_{L-1}, ..., \bar{h}_{l+1})\right)
\end{equation}
The TD network satisfies two conditions. First, we restrict $\bar{h}_l$ for every $l$ such that $h_l$ and $\bar{h}_l$ will have the same size (the same number of neurons). Hence, we can define pairs of corresponding neurons by assigning for each BU neuron $h_{l, i}$ in layer $l$, its 'counter neuron' to be the TD neuron $\bar{h}_{l, i}$. We also use the following notation for simplicity: $\bar{\bar{h}} \coloneqq h$. Additionally, we restrict $\bar{f}_l$ to have the same connectivity structure as $f_l$, but with the opposite direction: each pair of TD neurons is linked if and only if a link exists between their corresponding BU counter neurons. For example, given a fully connected layer $h_l = f_l(h_{l-1}) = W_{l} h_{l-1}$, the corresponding TD layer $\bar{f}_l$ is defined to be also a fully connected layer $\bar{h}_{l-1} = \bar{f}_l(\bar{h}_{l}) = \bar{W}_{l} \bar{h}_{l}$ such that the shape of the TD weights matrix $\bar{W}_l$ is equal to the shape of the transposed BU weights matrix ${W_l}^T$.

\subsection{Activation functions and biases}
\label{section - activation functions}
The activation functions $\sigma$, $\bar{\sigma}$, may be any element-wise functions. In this work, we focus on two functions. The first is ReLU which is commonly used for neural networks $ReLU(x) \coloneqq x \cdot I_{\{x > 0\}}$.


The second is the Gated-Linear-Unit (GaLU), which exploits the lateral connectivity between the BU and TD streams by gating neurons' activity according to the activation of their counter neurons.
\begin{equation}
\label{Couter Gating Def}
    GaLU(x) \coloneqq GaLU(x, \bar{x}) \coloneqq x \cdot I_{\{\bar{x} > 0\}} = \begin{cases} x & \bar{x} > 0 \\ 0 & \bar{x} \leq 0\end{cases}
\end{equation}
Where $\bar{x}$ is the counter neuron of $x$ (either a BU or a TD neuron), and $I$ is an indicator function.

Using GaLU introduces bidirectional lateral connectivity between the BU and TD networks by temporarily turning off neurons based on the values of their counter neurons. As a result, each network can effectively guide its counterpart to operate on a specific partial sub-network. 

In this paper, bias terms are omitted to simplify the model. Nevertheless, biases can be implicitly expressed using the above notations by having additional neurons and weights, as commonly practiced \cite{lee2015difference, ahmad2020gait}. In addition, we allow two modes of biases. The first is the standard bias mechanism, in which biases contribute to the output. The second mode is 'bias-blocking' \cite{akrout2019deep} in which all bias terms are zeroed.

\section{Counter-Hebbian learning}
\label{section CH learning}
\begin{figure}[ht]
  \centering
  \includegraphics[width=0.6\linewidth]{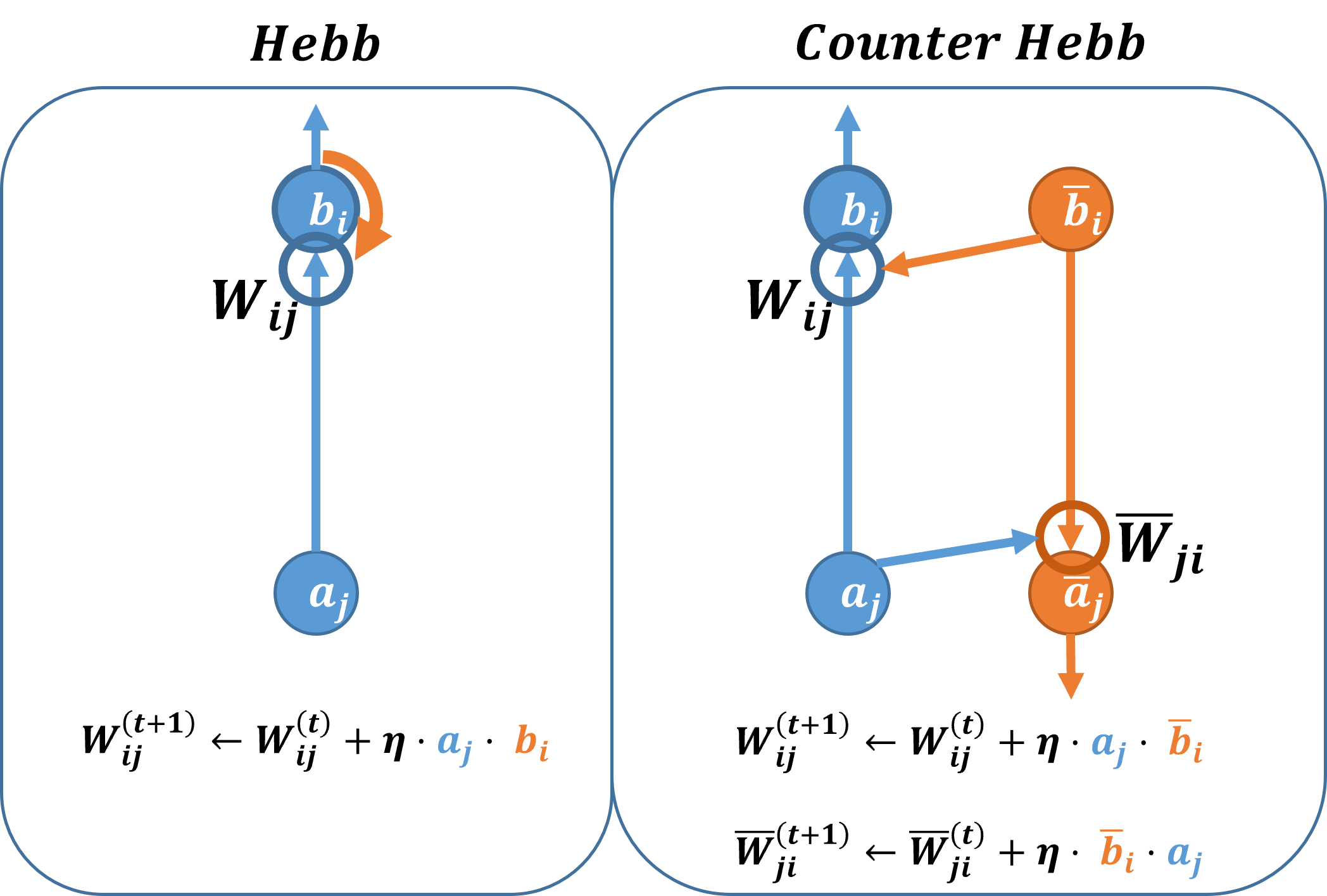}
  \caption{The Counter-Hebb update rule in comparison with the classical Hebb rule. The classical Hebbrule (on the left), with a focus on a single upstream synapse $W_{ij}$ (outlined by a circle), connecting a pre-synaptic neuron $a_j$ with a post-synaptic neuron $b_i$. The synapse $W_{ij}$ is updated based on the activity of both associated neurons $a_j$ and $b_i$. While neuron $a_j$ is directly associated with the synapse $W_{ij}$, neuron $b_i$ is assumed to transmit its information through propagation down the dendritic tree to synapse $W_{ij}$ (orange arrow). In contrast, the Counter-Hebb update rule, (on the right), relies on a contribution from the counterpart downstream (marked in orange),  mediated via lateral connections. Compared with the Hebb rule, the signal from  $a_j$ is combined with the signal from neuron $\bar{b}_i$ rather than neuron $b_i$. Notably, the resulting Counter-Hebb rule naturally applies an identical update to both $W_{i j}$ and its counter synapse $\bar{W}_{j i}$.}
  \label{fig: Update Rule}
\end{figure}

In this section, we formulate the Counter-Hebb learning. The proposed Counter-Hebb rule updates a given synapse based on the activities of its pre-synaptic neuron and its post-synaptic counter neuron. Consider a given weights matrix $W$ such that $b = W a$. The $i,j$-th entry in that matrix, $W^{(t)}_{i j}$, represents the strength of the synapse connecting the pre-synaptic neuron $a_j$ to the post-synaptic neuron $b_i$ at time $t$. Then the update rule is: 
\begin{equation}
\label{equation: Update Rule}
    \Delta W^{(t+1)}_{i j} \coloneqq W^{(t+1)}_{i j} - W^{(t)}_{i j} = \eta \cdot a_j \cdot \bar{b}_i
\end{equation}
where $\eta$ is the learning rate, and $\bar{b}_i$ is the counter neuron of $b_i$. This rule applies to all weights including both up-streams and down-streams, updating both $W$ and $\bar{W}$ identically, see Figure ~\ref{fig: Update Rule}.

There is a close connection between this rule and the classic Hebb rule. In both cases, the brain strengthens synapses (weights) between neurons that are co-activated, and the modifications of each synapse are determined entirely by the activation values of neurons in the network associated directly with the changing synapse. However, The difference lies in the neurons connected to that synapse. Classic Hebbian proposes that the forward-firing of a post-synaptic neuron also propagates backward to the synapse. Thereby, synapse strength increases when a pre-synaptic neuron's firing is often followed by the firing of the post-synaptic neuron within a defined time interval \cite{magee1997synaptically, hebb2005organization}. In contrast, Counter-Hebb modification does not depend on the cell's firing propagating back to its dendrites,  but suggests a contribution from the counter post-synaptic neuron via lateral connection. Similar to Hebb's rule, the resulting synaptic modification also depends on the coincidence of two firing neurons, but the post-synaptic cell is replaced by its counterpart. See Figure ~\ref{fig: Update Rule} for a visual illustration of the Counter-Hebb update compared with the classic Hebb. 

Therefore, The Counter-Hebb rule modifies the classical Hebb by incorporating feedback streams that can carry error information into the learning process. Empirical findings e.g. from CA1 hippocampal cells support the feasibility of synaptic plasticity that depends on the coincidence of two signals, from feedforward and feedback sources \cite{markov2014anatomy, cornford2019dendritic}.

\subsection{The Counter-Hebb learning algorithm and backpropagation}
\label{section: learning algorithm}

This section presents the full Counter-Hebbian (CH) learning algorithm. The learning algorithm is described in Algorithm ~\ref{alg: CH learning}. Similar to the backpropagation algorithm, the CH algorithm involves a single forward pass performed by the BU network to compute predictions from an input signal. Subsequently, a single backward pass is conducted using the TD network to propagate error information, and the weights are updated according to the CH update rule.

\begin{algorithm}[tb]
   \caption{Counter-Hebb Learning}
   \label{alg: CH learning}
\begin{algorithmic}[1]
   \STATE {\bfseries Input:} data $x$, ground truth label $\Tilde{y}$
   \STATE {\bfseries Forward}: $y = BU(x)$
   \STATE {\bfseries Compute error}: $e = error(y, \Tilde{y})$
   \STATE {\bfseries Backward}: $\bar{x} = TD(e)$
   \STATE {\bfseries Counter-Hebb Update}: update $W$ and $\bar{W}$
\end{algorithmic}
\end{algorithm}

A special case occurs when the BU and TD networks have symmetric weights, (identical values). While symmetric BU and TD weights might, at first, seem unrealistic in the brain, symmetry is actually implicitly encouraged by the CH update. The CH update naturally applies an identical update to both the BU and the TD weights, see Figure \ref{fig: Update Rule}. Therefore, as training progresses, assuming close-to-zero initial weights (a common practice), the BU and TD weights gradually become more symmetric, as the value of the weights will be dominant by the values of the updates. Moreover, at any point during the training, if the BU and TD weights are symmetric, they will maintain this symmetry during the entire learning.

Given symmetric BU and TD weights, under the following standard conditions: 1) The BU network uses ReLU non-linearity 2) The error function computes the negative gradients of a loss function $L$ with respect to the BU output, for example, $error(y, \Tilde{y}) = \Tilde{y} - y$ for Mean Squared Error loss 3) The TD network uses GaLU non-linearity and bias-blocking mode (see Section ~\ref{section - activation functions}). Then the TD backward step in Algorithm ~\ref{alg: CH learning} is mathematically equivalent to the backward computation of the BP algorithm \cite{rumelhart1986learning}. As a result, in this configuration, Counter-Hebb learning effectively replicates the exact BP update, performing similarly to BP and preserving its mathematical properties. Moreover, relaxing the symmetry constraint under the above conditions results in a learning algorithm that approximates BP in the non-symmetric case. For a detailed explanation of this equivalence and approximation see Appendix ~\ref{app: CH learning - BP}. Like some previous models, the CH has the desired property, as a biological model, of locality: the synaptic modifications are determined entirely by the activation values of neurons directly associated with the synapse. 

\section{Instruction-based learning}  
\label{section: mtl}
In the previous section, we described how the TD network can be used for learning a pure BU model. In this section, we describe how the model performs visual guidance. The TD network in our model can guide the BU network to perform multiple tasks by selecting a sub-network for each learned task (where sub-networks can overlap). In this setting, the objective is to predict an output $y$ given an input $x$ and a task $t$. To accommodate this, the model has one additional head, resulting in two heads: a prediction head, and an instruction head. Each head consists of two parts: one for the BU network and the other for the TD network, preserving the symmetrical structure and lateral connectivity of the BU-TD core, see Appendix Fig ~\ref{fig: heads} for an illustration of the heads' structure.  

\begin{figure}[ht]
  \centering
  \includegraphics[width=0.8\linewidth]{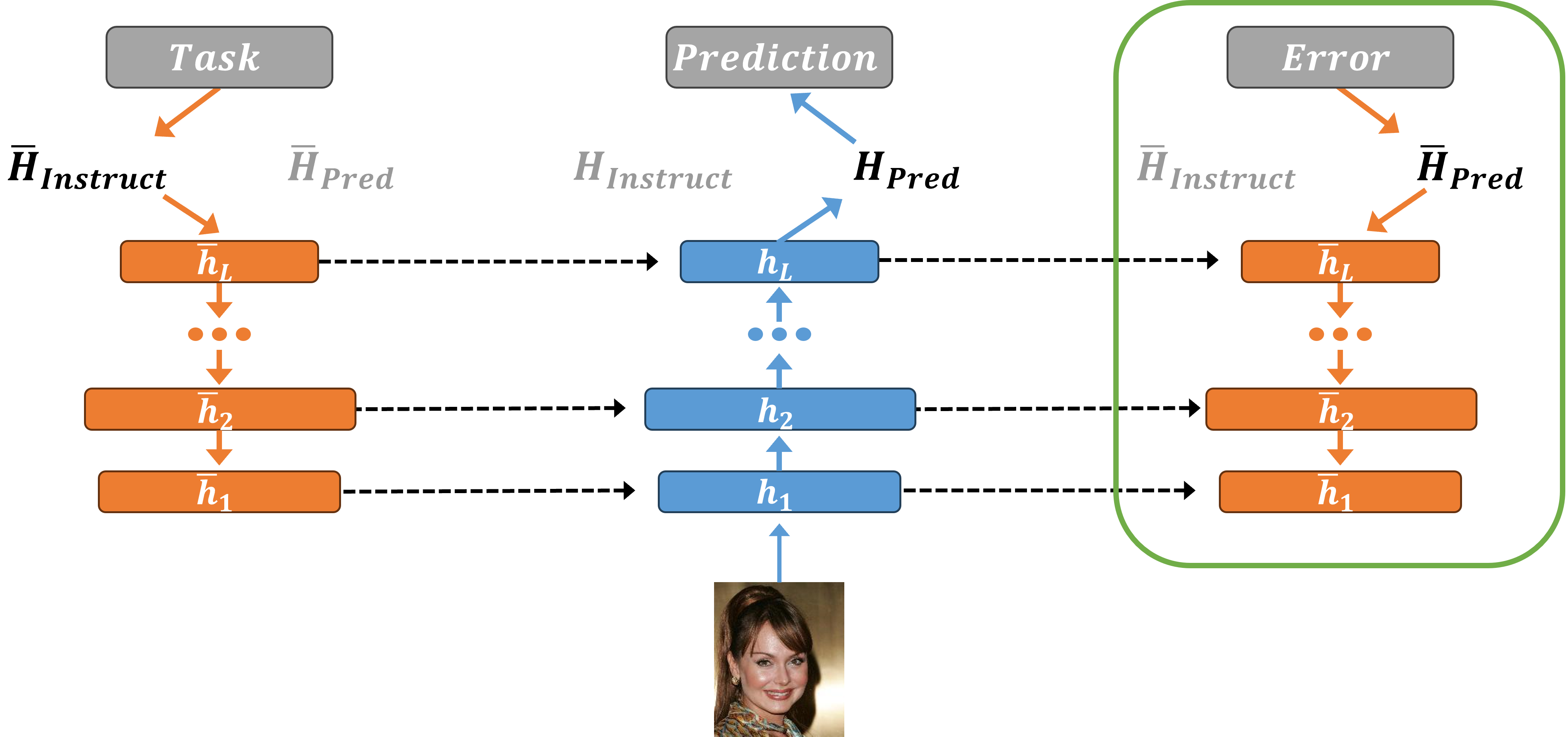}
  \caption{The instruction-based learning algorithm. The three columns represent three passes of our model (left to right): $TD \rightarrow BU \rightarrow TD$, where the first two passes provide a prediction output given an image and a task, and the last TD pass (in green frame) is used for learning. In \textit{inference}, The BU visual process is guided by the TD network according to the given task. More specifically, The TD network propagates downward instruction signals followed by a guided BU process of the input image to compute predictions. By applying ReLU non-linearity, the input task selectively activates a subset of neurons (i.e. non-zero values), composing a sub-network within the full network. The BU network then processes an input image using a composition of ReLU and GaLU. The GaLU function (dashed arrows) gates the BU computation to operate only on the selected sub-network that was activated by the task. For \textit{learning}, the same TD network is then reused to propagate prediction error signals with GaLU exclusively (no ReLU). Finally, the 'Counter-Hebb' learning rule adjusts both networks' weights based on the activation values of their neurons. Therefore, in contrast with standard models, the entire computation, including the learning, is carried out by neurons in the network, and no additional computation is used for learning (e.g. backpropagation)}
  \label{fig: mtl algorithm}
\end{figure}

The prediction head $H_{pred}$, of one linear layer, is responsible for generating predictions and providing feedback, as mentioned in section ~\ref{section: the BU-TD model}. The instruction head, $H_{instruct}$, employs a 2-layer MLP for specifying the selected task, projecting instructional information to the visual space (and vice versa), similarly to instruction processing in VLMs. More specifically, the instruction head takes a task representation $t$ as input and maps it to the top-level TD layer $\bar{h}_L$. We use one-hot encoding for representing the tasks, however, more complex embedding can be explored, such as a projection from an LLM. Note that in our experiments, we allow only one head to participate in each pass of the model (either the prediction or instruction head), refer to Fig ~\ref{fig: mtl algorithm} for an illustration of how the two heads are utilized in learning instruction-based models.

The instruction-based learning algorithm is shown in Algorithm ~\ref{alg: mtl}. This algorithm consists of two passes for prediction (a TD followed by BU) followed by an additional TD pass for the learning, thereby extending Algorithm ~\ref{alg: CH learning} with one additional step of instruction processing that selects a task-specific sub-network. Given a task $t$, the TD instruction head is used to propagate the task representation along the TD network. Since each task activates different patterns, the activated neurons (i.e., with activation value larger than $0$) define a task-dependent sub-network. By running the BU network with GaLU activation, the BU computation is gated to propagate the input $x$ along the corresponding BU sub-network. In this manner, the resulting algorithm learns for each task a different predictor which is conditioned on the task, resembling a modular architecture where different modules are dedicated to each task. See Fig \ref{fig: mtl algorithm} for a visualization of this guided process.

\begin{algorithm}[tb]
   \caption{Instruction-Based Learning}
   \label{alg: mtl}
\begin{algorithmic}[1]
   \STATE {\bfseries Input:} data $x$, task $t$, ground truth label $\Tilde{y}$
   \STATE {\bfseries Top-Down}: $\bar{x} = TD(t; \hspace{0.2cm} \bar{H}_{instruct}; \hspace{0.2cm} \bar{\sigma} = ReLU)$ 
   \label{mtl algo - td task process}
   \STATE {\bfseries Bottom-Up}: $y = BU(x; \hspace{0.2cm} H_{pred}; \hspace{0.2cm} \sigma = GaLU \circ ReLU)$ 
   \label{mtl algo - bu process}
   \STATE {\bfseries Compute error}: $e = error(y, \Tilde{y})$
   \STATE {\bfseries Backward}: $\bar{x} = TD(e; \hspace{0.2cm} \bar{H}_{pred}; \hspace{0.2cm} \bar{\sigma} = GaLU)$ 
   \STATE {\bfseries Counter-Hebb Update}: update $W$ and $\bar{W}$
\end{algorithmic}
\end{algorithm}

Extending upon the results of section ~\ref{section: learning algorithm}, under the constraint of symmetric BU and TD weights, surprisingly, both BP and CH learning yield identical updates in this guided learning framework, see Appendix ~\ref{app: CH learning - BP} for more details. This equivalence provides mathematical guarantees to learning guided visual processing using a single TD network for both guidance and learning. Furthermore, symmetric weights have computational advantages. It enables extending standard BU architectures to instruction-based models without any additional parameters. Given a BU network, a complementary symmetric TD network can be constructed, sharing the same BU parameters. This TD network can guide the BU process of the original network to perform a given instruction.

\section{Empirical results}

In this section, we evaluate our BU-TD model, learned via Counter-Hebbian learning, in two settings: 1) unguided visual processing, to show that CH learning is capable of learning vision models 2) guided visual processing, to evaluate the ability of our model to guide the visual process according to instructions. Our goal is not to improve upon state-of-the-art models, but to show that the model, with a single top-down pathway for both error and instruction propagation, is comparable with current AI models, and capable of performing well two different functions: learning and directing attention.

\subsection{Unguided visual processing}
In the unguided experiments, we evaluate the performance of the Counter-Hebb learning on standard image classification benchmarks: MNIST \cite{lecun1998gradient}, Fashion-MNIST \cite{xiao2017fashion}, and CIFAR10 \cite{krizhevsky2009learning}. We followed the same experiments as \citet{bozkurt2024correlative} and used two-layer fully connected networks, with a hidden layer of size 500 for both MNIST and Fashion-MNIST datasets and size 1,000 for CIFAR10. Further details including the full set of hyperparameters can be found in Appendix ~\ref{appendix: stl}. We compare CH learning using the Cross-Entropy loss with backpropagation and other biological learning methods. 

We examine two settings of CH learning, one where the BU and TD weights are initialized with symmetrical values, denoted as 'Sym Init', and the other where the weights are initialized differently, referred to as 'Asym Init'. The results, shown in table ~\ref{table: STL results}, empirically validate that CH learning is equivalent to BP in the symmetric case, and approximates BP in the asymmetric case. Moreover, CH learning achieves comparable or superior performance compared with other biological methods. We further show the robustness of CH on other architectures and settings, such as convolutional networks, loss functions, and regularization. The results and additional information regarding these experiments can be found in Appendix ~\ref{appendix: stl}.

\begin{table}[t]
    \caption{Unguided learning results: mean and standard deviation of the test accuracy (in percentages) across 10 runs. The proposed CH learning algorithm is compared with BP and other biological state-of-the-art methods. The baseline results were taken from \citet{bozkurt2024correlative}.}
    \label{table: STL results}
    \begin{center}
    \begin{tabular}{llll}
        \toprule
        Method     & MNIST  & Fashion MNIST     & CIFAR10 \\
        \midrule
        CIM [\citeyear{bozkurt2024correlative}] & 97.71 $\pm$ 0.1 & 88.14 $\pm$ 0.3 & 51.86 $\pm$ 0.3 \\
        EP [\citeyear{scellier2017equilibrium}] & 97.61 $\pm$ 0.1 & 88.06 $\pm$ 0.7 & 49.28 $\pm$ 0.5 \\
        CSM [\citeyear{qin2021contrastive}] & 98.08 $\pm$ 0.1 & 88.73 $\pm$ 0.2 & 40.79 \\
        PC [\citeyear{whittington2017approximation}] & 98.17 $\pm$ 0.2 & 89.31 $\pm$ 0.4 & - \\
        PC-Nudge [\citeyear{millidge2022backpropagation}] & 97.71 $\pm$ 0.1 & 88.49 $\pm$ 0.3 & 48.58 $\pm$ 0.7 \\
        FA [\citeyear{lillicrap2016random}] & 97.95 $\pm$ 0.08 & 88.38 $\pm$ 0.9 & 52.37 $\pm$ 0.4 \\
        BP & 98.27 $\pm$ 0.03 & 89.41 $\pm$ 0.2 & 53.96 $\pm$ 0.3 \\
        \midrule
        BP (ours) & 98.33 $\pm$ 0.04 & 89.94 $\pm$ 0.2 & 55.47 $\pm$ 0.3 \\
        CH Sym Init & 98.34 $\pm$ 0.06 & 89.99 $\pm$ 0.2 & 55.54 $\pm$ 0.3 \\
        CH Asym Init & 98.17 $\pm$ 0.06 & 89.27 $\pm$ 0.1 & 54.28 $\pm$ 0.2 \\
        \bottomrule
      \end{tabular}
    \end{center}
\end{table}

\subsection{Guided visual processing}

In the guided experiments, we evaluate our model on two common multi-task learning (MTL) benchmarks. Since current biological methods are not capable of guided processing, we compare CH with state-of-the-art non-biological optimization methods as reported by \citet{kurin2022defense}, replicating their setup and use their reported results. 

\textbf{The Multi-MNIST} dataset contains images of two overlaid digits, where the task indicates whether to classify the left or the right digit. Similar to the baselines, our BU network employs a simple architecture composed of two convolutional layers followed by a single fully-connected layer, with ReLU non-linearity, along with an additional fully-connected linear layer as the decoder (prediction head). To adapt this architecture to the BU-TD structure, we replace all max-pool layers with strided convolution layers, that perform a similar function as proposed by \cite{ayachi2020strided}. Since the BU-TD model uses only sparse sub-networks within the full network, we increased the number of channels in each convolution layer, however, the actual network size is effectively smaller compared with the baselines, see Appendix ~\ref{appendix: sub-networks analysis} for an analysis of the actual size of the sub-networks. Unlike the baselines, which employ a separate decoder (prediction head) for each task, our BU-TD model can use a single decoder for all tasks. We provide additional experiments in the guided regime including using a single decoder as well as weak symmetry and varied network sizes in Appendix ~\ref{appendix: exp}.

\textbf{The CelebA} dataset is a more challenging large-scale benchmark, comprising head shots of celebrities, along with the indication of the presence or absence of 40 different attributes. Each task is a binary classification problem for an attribute. As done in previous work \cite{kurin2022defense}, we employ a ResNet-18 \cite{he2016deep} architecture (without the final layer) with batch normalization layers \cite{ioffe2015batch}, and a linear decoder. Additionally, we remove the last average pooling layer to support the symmetric BU-TD structure. Further details including the exact architectures and hyper-parameters can be found in Appendix ~\ref{appendix: exp}.

\begin{table}[t]
    \caption{Guided processing results: mean and 95\% confidence interval of the avg. task test accuracy (in percentages) across 10 runs for \textit{Multi-MNIST} and 5 runs for \textit{CelebA}. The proposed CH learning algorithm is compared with non-biological state-of-the-art methods, as reported in \citeauthor{kurin2022defense}.}
    \label{table: MTL results}
    \begin{center}
    \begin{tabular}{lll}
    \toprule
    Method     & Multi-MNIST     & CelebA \\
    \midrule
    Unit. Scal. [\citeyear{kurin2022defense}] & 94.76 $\pm$ 0.44 &  90.90 $\pm$ 0.08 \\
    IMTL [\citeyear{liu2021towards}] & 94.87 $\pm$ 0.25 &  90.93 $\pm$ 0.08 \\
    MGDA [\citeyear{sener2018multi}] & 94.78 $\pm$ 0.20 & 90.22 $\pm$ 0.10 \\
    GradDrop [\citeyear{chen2020just}] & 93.47 $\pm$ 1.30 & 90.98 $\pm$ 0.03 \\
    PCGrad [\citeyear{yu2020gradient}] & 94.79 $\pm$ 0.36 & 90.93 $\pm$ 0.11 \\
    RLW Diri. [\citeyear{lin2021closer}] & 94.30 $\pm$ 0.30 & 90.99 $\pm$ 0.08 \\
    RLW Norm. [\citeyear{lin2021closer}] & 93.99 $\pm$ 0.89 & 90.95 $\pm$ 0.10 \\
    \midrule
    CH Asym Init & 88.92 $\pm$ 2.15 & 79.25 $\pm$ 1.63  \\
    CH Sym Init & 94.20 $\pm$ 0.30 & 89.69 $\pm$ 0.12 \\
    \bottomrule
    \end{tabular}
    \end{center}
\end{table}

The results, presented in Table ~\ref{table: MTL results}, show that the proposed model successfully incorporates the two different TD functions, directing attention, and learning. The BU-TD model can achieve competitive performance compared with leading non-biological state-of-the-art methods. The proposed method may offer additional useful computational properties, such as compactness, see Appendix ~\ref{appendix: sub-networks analysis}.  


\section{Limitations}

There are two directions that should be improved in the current model, one regarding performance and the second concerning biological aspects. In the asymmetric case, we observe greater performance gaps as task complexity increases, similarly as observed by other biological methods \cite{ernoult2022towards}. However, the learning parameters, including the weight initialization, were carefully optimized for symmetric weights over the years. Therefore, examining the effects of tuning parameters in the asymmetric case could improve performance. On the biological side, the plausibility of the model should be further explored. The main issue we identify (in ours and other models) is that propagating error signals along the TD stream requires the representation of both positive and negative values in neuronal activity \cite{lillicrap2020backpropagation}. Following initial work, we suggest that this could be obtained by ‘on’ and ‘off’ channels \cite{ringach2004mapping}, where negative values in the ‘on’ channel are represented by positive values in the complementary ‘off’ channel.

\section{Discussion}

In this paper, we proposed the first biologically-motivated learning model for instructed visual processing. Similar to the visual cortex, it uses a bottom-up (BU) top-down (TD) structure, which, unlike previous learning models, uses the TD stream in ongoing visual processing by directing attention, e.g. to tasks and locations of interest. 

Modeling learning in guided vision is challenging, since in guided vision, the prediction of the model depends on both the BU processing of the image and the task selected by a top-down instruction. The error signal needs, therefore, to propagate through both the BU and TD pathways, and at the same time the network is required to preserve the neural activations that existed during the prediction phase, since they determine the required changes in synaptic weights (Fig. ~\ref{fig: Update Rule}). These requirements place significant constraints on the structure of the model network, however, our model meets the requirements and, as supported by experiments, succeeds in learning guided vision for multiple tasks. Since the cortex performs similar guided vision, the proposed model may suggest a sketch model for the combination of the BU and TD streams in the visual cortex. Furthermore, the model shares a similar general structure with VLMs, in the sense of using two parallel streams, a visual one together with a more cognitive one. Since the human brain excels at combining visual and cognitive information in visual perception, the combination of instructed VLMs with principles from human BU-TD processing can offer a promising direction for future studies. 

The model suggests a Counter-Hebbian learning process in addition to the classical Hebb rule, where synapses are modified by combining a pre-synaptic signal with a signal coming from the appropriate counter stream. It may be possible to test the existence of CH learning biologically by the controlled activation of selected layers. For example, cortical layer 3B receives feedforward connections from layer 4 while feedback connections arrive to layer 2/3A \cite{markov2014anatomy}. The counter-Hebbian model predicts that it may be possible to modify the forward synapses from layer 4 to layer 3B by simultaneous activation of the two inputs.

\bibliographystyle{unsrtnat}
\bibliography{main}

\begin{thebibliography}{72}
\providecommand{\natexlab}[1]{#1}
\providecommand{\url}[1]{\texttt{#1}}
\expandafter\ifx\csname urlstyle\endcsname\relax
  \providecommand{\doi}[1]{doi: #1}\else
  \providecommand{\doi}{doi: \begingroup \urlstyle{rm}\Url}\fi

\bibitem[Ernoult et~al.(2022)Ernoult, Normandin, Moudgil, Spinney, Belilovsky, Rish, Richards, and Bengio]{ernoult2022towards}
Maxence~M Ernoult, Fabrice Normandin, Abhinav Moudgil, Sean Spinney, Eugene Belilovsky, Irina Rish, Blake Richards, and Yoshua Bengio.
\newblock Towards scaling difference target propagation by learning backprop targets.
\newblock In \emph{International Conference on Machine Learning}, pages 5968--5987. PMLR, 2022.

\bibitem[Bozkurt et~al.(2024)Bozkurt, Pehlevan, and Erdogan]{bozkurt2024correlative}
Bariscan Bozkurt, Cengiz Pehlevan, and Alper Erdogan.
\newblock Correlative information maximization: A biologically plausible approach to supervised deep neural networks without weight symmetry.
\newblock \emph{Advances in Neural Information Processing Systems}, 36, 2024.

\bibitem[Whittington and Bogacz(2019)]{whittington2019theories}
James~CR Whittington and Rafal Bogacz.
\newblock Theories of error back-propagation in the brain.
\newblock \emph{Trends in cognitive sciences}, 23\penalty0 (3):\penalty0 235--250, 2019.

\bibitem[Lillicrap et~al.(2020)Lillicrap, Santoro, Marris, Akerman, and Hinton]{lillicrap2020backpropagation}
Timothy~P Lillicrap, Adam Santoro, Luke Marris, Colin~J Akerman, and Geoffrey Hinton.
\newblock Backpropagation and the brain.
\newblock \emph{Nature Reviews Neuroscience}, 21\penalty0 (6):\penalty0 335--346, 2020.

\bibitem[Song et~al.(2021)Song, Xu, and Lafferty]{song2021convergence}
Ganlin Song, Ruitu Xu, and John Lafferty.
\newblock Convergence and alignment of gradient descent with random backpropagation weights.
\newblock \emph{Advances in Neural Information Processing Systems}, 34:\penalty0 19888--19898, 2021.

\bibitem[Manita et~al.(2015)Manita, Suzuki, Homma, Matsumoto, Odagawa, Yamada, Ota, Matsubara, Inutsuka, Sato, et~al.]{manita2015top}
Satoshi Manita, Takayuki Suzuki, Chihiro Homma, Takashi Matsumoto, Maya Odagawa, Kazuyuki Yamada, Keisuke Ota, Chie Matsubara, Ayumu Inutsuka, Masaaki Sato, et~al.
\newblock A top-down cortical circuit for accurate sensory perception.
\newblock \emph{Neuron}, 86\penalty0 (5):\penalty0 1304--1316, 2015.

\bibitem[Gilbert and Li(2013)]{gilbert2013top}
Charles~D Gilbert and Wu~Li.
\newblock Top-down influences on visual processing.
\newblock \emph{Nature Reviews Neuroscience}, 14\penalty0 (5):\penalty0 350--363, 2013.

\bibitem[Harel et~al.(2014)Harel, Kravitz, and Baker]{harel2014task}
Assaf Harel, Dwight~J Kravitz, and Chris~I Baker.
\newblock Task context impacts visual object processing differentially across the cortex.
\newblock \emph{Proceedings of the National Academy of Sciences}, 111\penalty0 (10):\penalty0 E962--E971, 2014.

\bibitem[Zagha(2020)]{zagha2020shaping}
Edward Zagha.
\newblock Shaping the cortical landscape: Functions and mechanisms of top-down cortical feedback pathways.
\newblock \emph{Frontiers in Systems Neuroscience}, 14:\penalty0 33, 2020.

\bibitem[Kreiman and Serre(2020)]{kreiman2020beyond}
Gabriel Kreiman and Thomas Serre.
\newblock Beyond the feedforward sweep: feedback computations in the visual cortex.
\newblock \emph{Annals of the New York Academy of Sciences}, 1464\penalty0 (1):\penalty0 222--241, 2020.

\bibitem[Hebb(2005)]{hebb2005organization}
Donald~Olding Hebb.
\newblock \emph{The organization of behavior: A neuropsychological theory}.
\newblock Psychology Press, 2005.

\bibitem[Rumelhart et~al.(1986)Rumelhart, Hinton, and Williams]{rumelhart1986learning}
David~E Rumelhart, Geoffrey~E Hinton, and Ronald~J Williams.
\newblock Learning representations by back-propagating errors.
\newblock \emph{nature}, 323\penalty0 (6088):\penalty0 533--536, 1986.

\bibitem[Huang et~al.(2023)Huang, Zhang, Jiang, Qiu, and Lu]{huang2023visual}
Jiaxing Huang, Jingyi Zhang, Kai Jiang, Han Qiu, and Shijian Lu.
\newblock Visual instruction tuning towards general-purpose multimodal model: A survey.
\newblock \emph{arXiv preprint arXiv:2312.16602}, 2023.

\bibitem[Liu et~al.(2024)Liu, Li, Wu, and Lee]{liu2024visual}
Haotian Liu, Chunyuan Li, Qingyang Wu, and Yong~Jae Lee.
\newblock Visual instruction tuning.
\newblock \emph{Advances in neural information processing systems}, 36, 2024.

\bibitem[Yamins and DiCarlo(2016)]{yamins2016using}
Daniel~LK Yamins and James~J DiCarlo.
\newblock Using goal-driven deep learning models to understand sensory cortex.
\newblock \emph{Nature neuroscience}, 19\penalty0 (3):\penalty0 356--365, 2016.

\bibitem[Bowers(2017)]{bowers2017parallel}
Jeffrey~S Bowers.
\newblock Parallel distributed processing theory in the age of deep networks.
\newblock \emph{Trends in cognitive sciences}, 21\penalty0 (12):\penalty0 950--961, 2017.

\bibitem[Yildirim et~al.(2019)Yildirim, Wu, Kanwisher, and Tenenbaum]{yildirim2019integrative}
Ilker Yildirim, Jiajun Wu, Nancy Kanwisher, and Joshua Tenenbaum.
\newblock An integrative computational architecture for object-driven cortex.
\newblock \emph{Current opinion in neurobiology}, 55:\penalty0 73--81, 2019.

\bibitem[Scellier and Bengio(2017)]{scellier2017equilibrium}
Benjamin Scellier and Yoshua Bengio.
\newblock Equilibrium propagation: Bridging the gap between energy-based models and backpropagation.
\newblock \emph{Frontiers in computational neuroscience}, 11:\penalty0 24, 2017.

\bibitem[Ernoult et~al.(2019)Ernoult, Grollier, Querlioz, Bengio, and Scellier]{ernoult2019updates}
Maxence Ernoult, Julie Grollier, Damien Querlioz, Yoshua Bengio, and Benjamin Scellier.
\newblock Updates of equilibrium prop match gradients of backprop through time in an rnn with static input.
\newblock \emph{Advances in neural information processing systems}, 32, 2019.

\bibitem[Millidge et~al.(2020)Millidge, Tschantz, Buckley, and Seth]{millidge2020activation}
Beren Millidge, Alexander Tschantz, Christopher~L Buckley, and Anil Seth.
\newblock Activation relaxation: A local dynamical approximation to backpropagation in the brain.
\newblock \emph{arXiv preprint arXiv:2009.05359}, 2020.

\bibitem[Whittington and Bogacz(2017)]{whittington2017approximation}
James~CR Whittington and Rafal Bogacz.
\newblock An approximation of the error backpropagation algorithm in a predictive coding network with local hebbian synaptic plasticity.
\newblock \emph{Neural computation}, 29\penalty0 (5):\penalty0 1229--1262, 2017.

\bibitem[Millidge et~al.(2022{\natexlab{a}})Millidge, Tschantz, and Buckley]{millidge2022predictive}
Beren Millidge, Alexander Tschantz, and Christopher~L Buckley.
\newblock Predictive coding approximates backprop along arbitrary computation graphs.
\newblock \emph{Neural Computation}, 34\penalty0 (6):\penalty0 1329--1368, 2022{\natexlab{a}}.

\bibitem[Song et~al.(2020)Song, Lukasiewicz, Xu, and Bogacz]{song2020can}
Yuhang Song, Thomas Lukasiewicz, Zhenghua Xu, and Rafal Bogacz.
\newblock Can the brain do backpropagation?—exact implementation of backpropagation in predictive coding networks.
\newblock \emph{Advances in neural information processing systems}, 33:\penalty0 22566, 2020.

\bibitem[Salvatori et~al.(2022)Salvatori, Song, Xu, Lukasiewicz, and Bogacz]{salvatori2022reverse}
Tommaso Salvatori, Yuhang Song, Zhenghua Xu, Thomas Lukasiewicz, and Rafal Bogacz.
\newblock Reverse differentiation via predictive coding.
\newblock In \emph{Proceedings of the AAAI Conference on Artificial Intelligence}, volume~36, pages 8150--8158, 2022.

\bibitem[Rosenbaum(2022)]{rosenbaum2022relationship}
Robert Rosenbaum.
\newblock On the relationship between predictive coding and backpropagation.
\newblock \emph{Plos one}, 17\penalty0 (3):\penalty0 e0266102, 2022.

\bibitem[Golkar et~al.(2022)Golkar, Tesileanu, Bahroun, Sengupta, and Chklovskii]{golkar2022constrained}
Siavash Golkar, Tiberiu Tesileanu, Yanis Bahroun, Anirvan Sengupta, and Dmitri Chklovskii.
\newblock Constrained predictive coding as a biologically plausible model of the cortical hierarchy.
\newblock \emph{Advances in Neural Information Processing Systems}, 35:\penalty0 14155--14169, 2022.

\bibitem[Lillicrap et~al.(2016)Lillicrap, Cownden, Tweed, and Akerman]{lillicrap2016random}
Timothy~P Lillicrap, Daniel Cownden, Douglas~B Tweed, and Colin~J Akerman.
\newblock Random synaptic feedback weights support error backpropagation for deep learning.
\newblock \emph{Nature communications}, 7\penalty0 (1):\penalty0 1--10, 2016.

\bibitem[N{\o}kland(2016)]{nokland2016direct}
Arild N{\o}kland.
\newblock Direct feedback alignment provides learning in deep neural networks.
\newblock \emph{Advances in neural information processing systems}, 29, 2016.

\bibitem[Bengio(2014)]{bengio2014auto}
Yoshua Bengio.
\newblock How auto-encoders could provide credit assignment in deep networks via target propagation.
\newblock \emph{arXiv preprint arXiv:1407.7906}, 2014.

\bibitem[Lee et~al.(2015)Lee, Zhang, Fischer, and Bengio]{lee2015difference}
Dong-Hyun Lee, Saizheng Zhang, Asja Fischer, and Yoshua Bengio.
\newblock Difference target propagation.
\newblock In \emph{Joint european conference on machine learning and knowledge discovery in databases}, pages 498--515. Springer, 2015.

\bibitem[Meulemans et~al.(2020)Meulemans, Carzaniga, Suykens, Sacramento, and Grewe]{meulemans2020theoretical}
Alexander Meulemans, Francesco Carzaniga, Johan Suykens, Jo{\~a}o Sacramento, and Benjamin~F Grewe.
\newblock A theoretical framework for target propagation.
\newblock \emph{Advances in Neural Information Processing Systems}, 33:\penalty0 20024--20036, 2020.

\bibitem[Akrout et~al.(2019)Akrout, Wilson, Humphreys, Lillicrap, and Tweed]{akrout2019deep}
Mohamed Akrout, Collin Wilson, Peter~C Humphreys, Timothy Lillicrap, and Douglas Tweed.
\newblock Deep learning without weight transport.
\newblock \emph{arXiv preprint arXiv:1904.05391}, 2019.

\bibitem[Ahmad et~al.(2020)Ahmad, van Gerven, and Ambrogioni]{ahmad2020gait}
Nasir Ahmad, Marcel~AJ van Gerven, and Luca Ambrogioni.
\newblock Gait-prop: A biologically plausible learning rule derived from backpropagation of error.
\newblock \emph{arXiv preprint arXiv:2006.06438}, 2020.

\bibitem[Wen et~al.(2019)Wen, Yu, Yang, and Li]{wen2019goal}
Zhenfu Wen, Tianyou Yu, Xinbin Yang, and Yuanqing Li.
\newblock Goal-directed processing of naturalistic stimuli modulates large-scale functional connectivity.
\newblock \emph{Frontiers in neuroscience}, 12:\penalty0 1003, 2019.

\bibitem[Dehaene et~al.(2021)Dehaene, Lau, and Kouider]{dehaene2021consciousness}
Stanislas Dehaene, Hakwan Lau, and Sid Kouider.
\newblock What is consciousness, and could machines have it?
\newblock \emph{Robotics, AI, and Humanity: Science, Ethics, and Policy}, pages 43--56, 2021.

\bibitem[Goddard et~al.(2022)Goddard, Carlson, and Woolgar]{goddard2022spatial}
Erin Goddard, Thomas~A Carlson, and Alexandra Woolgar.
\newblock Spatial and feature-selective attention have distinct, interacting effects on population-level tuning.
\newblock \emph{Journal of cognitive neuroscience}, 34\penalty0 (2):\penalty0 290--312, 2022.

\bibitem[Shahdloo et~al.(2022)Shahdloo, {\c{C}}elik, Urgen, Gallant, and {\c{C}}ukur]{shahdloo2022task}
Mo~Shahdloo, Emin {\c{C}}elik, Burcu~A Urgen, Jack~L Gallant, and Tolga {\c{C}}ukur.
\newblock Task-dependent warping of semantic representations during search for visual action categories.
\newblock \emph{Journal of Neuroscience}, 42\penalty0 (35):\penalty0 6782--6799, 2022.

\bibitem[Tsotsos(2021)]{tsotsos2021computational}
John~K Tsotsos.
\newblock \emph{A computational perspective on visual attention}.
\newblock MIT Press, 2021.

\bibitem[Pang et~al.(2021)Pang, Li, Li, Li, Cao, and Lu]{pang2021tdaf}
Bo~Pang, Yizhuo Li, Jiefeng Li, Muchen Li, Hanwen Cao, and Cewu Lu.
\newblock Tdaf: Top-down attention framework for vision tasks.
\newblock In \emph{Proceedings of the AAAI Conference on Artificial Intelligence}, volume~35, pages 2384--2392, 2021.

\bibitem[Ullman et~al.(2023)Ullman, Assif, Strugatski, Vatashsky, Levi, Netanyahu, and Yaari]{ullman2023human}
Shimon Ullman, Liav Assif, Alona Strugatski, Ben-Zion Vatashsky, Hila Levi, Aviv Netanyahu, and Adam Yaari.
\newblock Human-like scene interpretation by a guided counterstream processing.
\newblock \emph{Proceedings of the National Academy of Sciences}, 120\penalty0 (40):\penalty0 e2211179120, 2023.

\bibitem[Bai et~al.(2023)Bai, Bai, Yang, Wang, Tan, Wang, Lin, Zhou, and Zhou]{bai2023qwen}
Jinze Bai, Shuai Bai, Shusheng Yang, Shijie Wang, Sinan Tan, Peng Wang, Junyang Lin, Chang Zhou, and Jingren Zhou.
\newblock Qwen-vl: A frontier large vision-language model with versatile abilities.
\newblock \emph{arXiv preprint arXiv:2308.12966}, 2023.

\bibitem[Zhu et~al.(2023)Zhu, Chen, Shen, Li, and Elhoseiny]{zhu2023minigpt}
Deyao Zhu, Jun Chen, Xiaoqian Shen, Xiang Li, and Mohamed Elhoseiny.
\newblock Minigpt-4: Enhancing vision-language understanding with advanced large language models.
\newblock \emph{arXiv preprint arXiv:2304.10592}, 2023.

\bibitem[Dai et~al.(2024)Dai, Li, Li, Tiong, Zhao, Wang, Li, Fung, and Hoi]{dai2024instructblip}
Wenliang Dai, Junnan Li, Dongxu Li, Anthony Meng~Huat Tiong, Junqi Zhao, Weisheng Wang, Boyang Li, Pascale~N Fung, and Steven Hoi.
\newblock Instructblip: Towards general-purpose vision-language models with instruction tuning.
\newblock \emph{Advances in Neural Information Processing Systems}, 36, 2024.

\bibitem[Magee and Johnston(1997)]{magee1997synaptically}
Jeffrey~C Magee and Daniel Johnston.
\newblock A synaptically controlled, associative signal for hebbian plasticity in hippocampal neurons.
\newblock \emph{Science}, 275\penalty0 (5297):\penalty0 209--213, 1997.

\bibitem[Markov et~al.(2014)Markov, Vezoli, Chameau, Falchier, Quilodran, Huissoud, Lamy, Misery, Giroud, Ullman, et~al.]{markov2014anatomy}
Nikola~T Markov, Julien Vezoli, Pascal Chameau, Arnaud Falchier, Ren{\'e} Quilodran, Cyril Huissoud, Camille Lamy, Pierre Misery, Pascale Giroud, Shimon Ullman, et~al.
\newblock Anatomy of hierarchy: feedforward and feedback pathways in macaque visual cortex.
\newblock \emph{Journal of Comparative Neurology}, 522\penalty0 (1):\penalty0 225--259, 2014.

\bibitem[Cornford et~al.(2019)Cornford, Mercier, Leite, Magloire, H{\"a}usser, and Kullmann]{cornford2019dendritic}
Jonathan~H Cornford, Marion~S Mercier, Marco Leite, Vincent Magloire, Michael H{\"a}usser, and Dimitri~M Kullmann.
\newblock Dendritic nmda receptors in parvalbumin neurons enable strong and stable neuronal assemblies.
\newblock \emph{Elife}, 8:\penalty0 e49872, 2019.

\bibitem[LeCun et~al.(1998)LeCun, Bottou, Bengio, and Haffner]{lecun1998gradient}
Yann LeCun, L{\'e}on Bottou, Yoshua Bengio, and Patrick Haffner.
\newblock Gradient-based learning applied to document recognition.
\newblock \emph{Proceedings of the IEEE}, 86\penalty0 (11):\penalty0 2278--2324, 1998.

\bibitem[Xiao et~al.(2017)Xiao, Rasul, and Vollgraf]{xiao2017fashion}
Han Xiao, Kashif Rasul, and Roland Vollgraf.
\newblock Fashion-mnist: a novel image dataset for benchmarking machine learning algorithms.
\newblock \emph{arXiv preprint arXiv:1708.07747}, 2017.

\bibitem[Krizhevsky et~al.(2009)Krizhevsky, Hinton, et~al.]{krizhevsky2009learning}
Alex Krizhevsky, Geoffrey Hinton, et~al.
\newblock Learning multiple layers of features from tiny images.
\newblock 2009.

\bibitem[Qin et~al.(2021)Qin, Mudur, and Pehlevan]{qin2021contrastive}
Shanshan Qin, Nayantara Mudur, and Cengiz Pehlevan.
\newblock Contrastive similarity matching for supervised learning.
\newblock \emph{Neural computation}, 33\penalty0 (5):\penalty0 1300--1328, 2021.

\bibitem[Millidge et~al.(2022{\natexlab{b}})Millidge, Song, Salvatori, Lukasiewicz, and Bogacz]{millidge2022backpropagation}
Beren Millidge, Yuhang Song, Tommaso Salvatori, Thomas Lukasiewicz, and Rafal Bogacz.
\newblock Backpropagation at the infinitesimal inference limit of energy-based models: Unifying predictive coding, equilibrium propagation, and contrastive hebbian learning.
\newblock \emph{arXiv preprint arXiv:2206.02629}, 2022{\natexlab{b}}.

\bibitem[Kurin et~al.(2022)Kurin, De~Palma, Kostrikov, Whiteson, and Mudigonda]{kurin2022defense}
Vitaly Kurin, Alessandro De~Palma, Ilya Kostrikov, Shimon Whiteson, and Pawan~K Mudigonda.
\newblock In defense of the unitary scalarization for deep multi-task learning.
\newblock \emph{Advances in Neural Information Processing Systems}, 35:\penalty0 12169--12183, 2022.

\bibitem[Ayachi et~al.(2020)Ayachi, Afif, Said, and Atri]{ayachi2020strided}
Riadh Ayachi, Mouna Afif, Yahia Said, and Mohamed Atri.
\newblock Strided convolution instead of max pooling for memory efficiency of convolutional neural networks.
\newblock In \emph{Proceedings of the 8th International Conference on Sciences of Electronics, Technologies of Information and Telecommunications (SETIT’18), Vol. 1}, pages 234--243. Springer, 2020.

\bibitem[He et~al.(2016)He, Zhang, Ren, and Sun]{he2016deep}
Kaiming He, Xiangyu Zhang, Shaoqing Ren, and Jian Sun.
\newblock Deep residual learning for image recognition.
\newblock In \emph{Proceedings of the IEEE conference on computer vision and pattern recognition}, pages 770--778, 2016.

\bibitem[Ioffe and Szegedy(2015)]{ioffe2015batch}
Sergey Ioffe and Christian Szegedy.
\newblock Batch normalization: Accelerating deep network training by reducing internal covariate shift.
\newblock In \emph{International conference on machine learning}, pages 448--456. PMLR, 2015.

\bibitem[Liu et~al.(2021)Liu, Li, Kuang, Xue, Chen, Yang, Liao, and Zhang]{liu2021towards}
Liyang Liu, Yi~Li, Zhanghui Kuang, J~Xue, Yimin Chen, Wenming Yang, Qingmin Liao, and Wayne Zhang.
\newblock Towards impartial multi-task learning.
\newblock iclr, 2021.

\bibitem[Sener and Koltun(2018)]{sener2018multi}
Ozan Sener and Vladlen Koltun.
\newblock Multi-task learning as multi-objective optimization.
\newblock \emph{Advances in neural information processing systems}, 31, 2018.

\bibitem[Chen et~al.(2020)Chen, Ngiam, Huang, Luong, Kretzschmar, Chai, and Anguelov]{chen2020just}
Zhao Chen, Jiquan Ngiam, Yanping Huang, Thang Luong, Henrik Kretzschmar, Yuning Chai, and Dragomir Anguelov.
\newblock Just pick a sign: Optimizing deep multitask models with gradient sign dropout.
\newblock \emph{Advances in Neural Information Processing Systems}, 33:\penalty0 2039--2050, 2020.

\bibitem[Yu et~al.(2020)Yu, Kumar, Gupta, Levine, Hausman, and Finn]{yu2020gradient}
Tianhe Yu, Saurabh Kumar, Abhishek Gupta, Sergey Levine, Karol Hausman, and Chelsea Finn.
\newblock Gradient surgery for multi-task learning.
\newblock \emph{Advances in Neural Information Processing Systems}, 33:\penalty0 5824--5836, 2020.

\bibitem[Lin et~al.(2021)Lin, Ye, and Zhang]{lin2021closer}
Baijiong Lin, Feiyang Ye, and Yu~Zhang.
\newblock A closer look at loss weighting in multi-task learning.
\newblock \emph{arXiv preprint arXiv:2111.10603}, 2021.

\bibitem[Ringach(2004)]{ringach2004mapping}
Dario~L Ringach.
\newblock Mapping receptive fields in primary visual cortex.
\newblock \emph{The Journal of Physiology}, 558\penalty0 (3):\penalty0 717--728, 2004.

\bibitem[LeCun et~al.(2015)LeCun, Bengio, and Hinton]{lecun2015deep}
Yann LeCun, Yoshua Bengio, and Geoffrey Hinton.
\newblock Deep learning.
\newblock \emph{nature}, 521\penalty0 (7553):\penalty0 436--444, 2015.

\bibitem[Frankle and Carbin(2018)]{frankle2018lottery}
Jonathan Frankle and Michael Carbin.
\newblock The lottery ticket hypothesis: Finding sparse, trainable neural networks.
\newblock \emph{arXiv preprint arXiv:1803.03635}, 2018.

\bibitem[Sabour et~al.(2017)Sabour, Frosst, and Hinton]{sabour2017dynamic}
Sara Sabour, Nicholas Frosst, and Geoffrey~E Hinton.
\newblock Dynamic routing between capsules.
\newblock \emph{Advances in neural information processing systems}, 30, 2017.

\bibitem[Liu et~al.(2015)Liu, Luo, Wang, and Tang]{liu2015deep}
Ziwei Liu, Ping Luo, Xiaogang Wang, and Xiaoou Tang.
\newblock Deep learning face attributes in the wild.
\newblock In \emph{Proceedings of the IEEE international conference on computer vision}, pages 3730--3738, 2015.

\bibitem[Ruder(2016)]{ruder2016overview}
Sebastian Ruder.
\newblock An overview of gradient descent optimization algorithms.
\newblock \emph{arXiv preprint arXiv:1609.04747}, 2016.

\bibitem[Malach et~al.(2020)Malach, Yehudai, Shalev-Schwartz, and Shamir]{malach2020proving}
Eran Malach, Gilad Yehudai, Shai Shalev-Schwartz, and Ohad Shamir.
\newblock Proving the lottery ticket hypothesis: Pruning is all you need.
\newblock In \emph{International Conference on Machine Learning}, pages 6682--6691. PMLR, 2020.

\bibitem[Chen et~al.(2021)Chen, Cheng, Wang, Gan, Liu, and Wang]{chen2021elastic}
Xiaohan Chen, Yu~Cheng, Shuohang Wang, Zhe Gan, Jingjing Liu, and Zhangyang Wang.
\newblock The elastic lottery ticket hypothesis.
\newblock \emph{Advances in Neural Information Processing Systems}, 34:\penalty0 26609--26621, 2021.

\bibitem[Morcos et~al.(2019)Morcos, Yu, Paganini, and Tian]{morcos2019one}
Ari Morcos, Haonan Yu, Michela Paganini, and Yuandong Tian.
\newblock One ticket to win them all: generalizing lottery ticket initializations across datasets and optimizers.
\newblock \emph{Advances in neural information processing systems}, 32, 2019.

\bibitem[Ramanujan et~al.(2020)Ramanujan, Wortsman, Kembhavi, Farhadi, and Rastegari]{ramanujan2020s}
Vivek Ramanujan, Mitchell Wortsman, Aniruddha Kembhavi, Ali Farhadi, and Mohammad Rastegari.
\newblock What's hidden in a randomly weighted neural network?
\newblock In \emph{Proceedings of the IEEE/CVF Conference on Computer Vision and Pattern Recognition}, pages 11893--11902, 2020.

\bibitem[Tanaka et~al.(2020)Tanaka, Kunin, Yamins, and Ganguli]{tanaka2020pruning}
Hidenori Tanaka, Daniel Kunin, Daniel~L Yamins, and Surya Ganguli.
\newblock Pruning neural networks without any data by iteratively conserving synaptic flow.
\newblock \emph{Advances in Neural Information Processing Systems}, 33:\penalty0 6377--6389, 2020.

\bibitem[Yu et~al.(2022)Yu, Serra, Ramalingam, and Zhe]{yu2022combinatorial}
Xin Yu, Thiago Serra, Srikumar Ramalingam, and Shandian Zhe.
\newblock The combinatorial brain surgeon: Pruning weights that cancel one another in neural networks.
\newblock In \emph{International Conference on Machine Learning}, pages 25668--25683. PMLR, 2022.

\end{thebibliography}


\appendix

\section{Appendix / supplemental material}

\subsection{Equivalent to the backpropagation algorithm}
\label{app: CH learning - BP}


In this section, we give a detailed explanation for the equivalence between the proposed Counter-Hebbian (CH) learning and Back-Propagation (BP), discussed in section ~\ref{section: learning algorithm}. BP is a key component of current learning algorithms for artificial neural networks, and modern deep learning models are typically optimized using end-to-end BP and a global loss function \cite{lecun2015deep}. BP is an efficient algorithm to compute gradients, designed especially for deep neural networks \cite{rumelhart1986learning}. The algorithm uses the chain rule to back-propagate error signals through the network, thus computing the gradients of a loss function $L$ with respect to all parameters through a single backward pass.

\subsubsection{Symmetric weights}

Given a feedforward BU network architecture, as described in section ~\ref{section: the BU-TD model}, Then, the BP backward pass propagates error signals $\delta$ through the network from the output layer according to:
\begin{equation}
    \label{back propagation equation}
    \delta_{l-1} \coloneqq \frac{\partial f_l}{\partial h_{l-1}} \delta_{l} = \sigma '(h_{l-1}) W_{l}^T \delta_{l}
\end{equation}
Where the initial $\delta$ values, that correspond to the output layer, are the derivative of the loss with respect to the output layer:  $\delta_{L} = \frac{\partial \text{Loss}}{\partial h_{L}}$. This construction of $\delta$ enables an easy way to compute the gradients with respect to each parameter: 
\begin{equation}
    \nabla \text{Loss}(W_l) = \frac{\partial L}{\partial W_l} = \delta_{l} h_{l-1}^T
\end{equation}

Given the following conditions: 
\begin{itemize}
    \item Symmetric BU and TD weights, $W = \bar{W}^T$
    \item The BU network uses ReLU non-linearity, $\sigma = ReLU$ 
    \item The error function computes the negative gradients of a loss function $L$ with respect to the BU output, $e = error(y, \Tilde{y}) = -\frac{\partial L(y, \Tilde{y})}{\partial y}$
    \item The TD network uses GaLU non-linearity, $\bar{\sigma} = GaLU$, and bias-blocking mode (see Section ~\ref{section - activation functions})
\end{itemize}

The TD process done in the backward step in Algorithm \ref{alg: CH learning} makes the exact same computation as BP at each layer:
\begin{equation}
    \Bar{h}_{l-1} \coloneqq \text{GaLU}(\bar{W}_{l} \bar{h}_{l}) = \sigma '(h_{l-1}) W_{l}^T \bar{h}_{l}
\end{equation}


This similarity is since the BU and TD weights are symmetric, i.e. $W_{l}^T = \bar{W}_{l}$, and the GaLU function effectively applies a product of $x$ with an indicator function which is exactly the gradient of the ReLU function, thus the GaLU operation is equivalent to multiplication with the derivatives of the BU ReLU function. 

Therefore, since the input to the TD network is the negative derivative of the loss function with respect to the output, the TD neurons have the same values as the BP signals, up to a different sign:
\begin{equation}
    \Bar{h}_{l} = -\delta_{l} 
\end{equation}

As a result, the update derived from our CH learning is equivalent to BP in this symmetric case and also performs a Gradient Descent (GD) update: 
\begin{equation}
    \Delta W_l = \eta \bar{h}_l h_{l-1}^T = -\eta \delta_l h_{l-1}^T = -\eta \nabla \text{Loss}(W_l)
\end{equation}

Moreover, when exploring the non-symmetric case, which has the same conditions as above but the symmetric constraint, we get that CH learning approximates the BP update as the learning progresses. 

\subsubsection{Asymmetric weights}

In the asymmetric case, the BU and TD weights are initialized with different values. Consider a BU weight matrix $W$ and its counter TD weights $\bar{W}$, where both weights were initialized i.i.d from a uniform distribution $U[-a ,a]$, where $a$ is a small positive scalar (a common practice). At each time step $t$ during the learning, the Counter-Hebb update applies a symmetrical update (up to the transposed dimensions of the matrices):
\begin{equation}
    \Delta W^{(t)} = \Delta \bar{W}^{(t)}
\end{equation}

Consequently, in each time step, the difference between the two weight matrices remains constant, and is determined by the initialization of the weight, thus is bounded:
\begin{equation*}
\begin{gathered}
     \left|W_{i j}^{(t)} -\bar{W}_{j i}^{(t)}\right| = \left|\left( W_{i j}^{(0)} + \sum_{k=1}^{t} \Delta W_{i j}^{(k)}\right) - \left(\bar{W}_{j i}^{(0)} + \sum_{k=1}^{t} \Delta \bar{W}_{j i}^{(k)}\right)\right| = \\
     = \left|W_{i j}^{(0)} -\bar{W}_{j i}^{(0)}\right| <= 2a
\end{gathered}
\end{equation*}

Notably, the weight initialization scheme is controlled, therefore the value $a$ can be controlled. Furthermore, a common belief is that high-magnitude weights of a trained network, are the weights that are important for the learned task. Pruning techniques have shown that those weights alone are sufficient for achieving results as good a full model consisting of all the weights \cite{frankle2018lottery}. Hence, focusing on a specific important weight (that has a high magnitude) $\left|W_{i j}^{(t)} \right| \gg 0$, then
\begin{equation*}
\begin{gathered}
    \left|\frac{W_{i j}^{(t)} -\bar{W}_{j i}^{(t)}}{W_{i j}^{(t)}}  \right| <= \left|\frac{2a}{W_{i j}^{(t)}}  \right| \approx 0
\end{gathered}
\end{equation*}

Therefore, as the training progresses, assuming close to zero weight initialization, the difference between the BU and TD weights will be negligible for the dominant BU weights that are important for the task. Consequently, as the training proceeds, the Counter-Hebb learning gradually pushes the BU and TD weights towards symmetry, and the CH update rule approximates the BP update. 

Moreover, we can make the weights converge to exact symmetry by adding a weight decay mechanism. Denoted the original update at time $t$ by $A(t)$, the new updates at time $t$ will be $\Delta W_l^{(t)} = A(t) - \lambda W^{(t)}$ and $\Delta \bar{W}_l^{(t) T} = A(t) - \lambda \bar{W}_l^{(t) T}$. Thus, 

\begin{equation*}
\begin{gathered}
    \left|W^{(t)} -\bar{W}^{(t) T} \right| = \left| (1-\lambda)^t W^{(0)} - (1-\lambda)^t \bar{W}^{(0) T}\right| \xrightarrow{t \rightarrow \infty}  0
\end{gathered}
\end{equation*}

Therefore, similarly to the results shown by \citet{akrout2019deep}, initializing the BU and TD weights with different values will converge to symmetric BU and TD weights, in which the CH learning is equivalent to backpropagation. Hence, in that non-symmetric case, the Counter-Hebb learning algorithm approximates the BP and approaches the exact BP.

\subsubsection{Guided visual processing}

Extending upon the above results to the guided learning framework, under the same constraints of the symmetric case, both BP and CH learning yield identical updates.

The guided learning algorithm consists of two passes for prediction, a TD pass followed by a BU pass. Hence, updating this model via BP requires computing the gradients of the loss function with respect to both the TD and BU computations. 

Notably, the first TD computation in the prediction phase, is connected to the final prediction only through the gating functions on the computation graph, see Figure ~\ref{fig: mtl algorithm}. Moreover, the gradients of this function with respect to the gate $\bar{x}$ are always zero. Therefore, this TD computation does not contribute any gradients to the prediction process:
\begin{equation}
    \nabla \text{Loss}(\bar{W}_l) = \frac{\partial L}{\partial \Bar{W}_l} = 0
\end{equation}

Focusing on the BU computation, given the constraint of symmetric BU and TD weights, the results from the non-guided scenario indicate that the last TD pass in our algorithm, used for error propagation, computes the exact backpropagation signals relative to the BU computation. 

Consequently, given the constraint of symmetric weights, for example, obtained by sharing the same weights across the two streams, the BP algorithm actually updates both the BU and TD weights according only to the gradients of the BU pass. Hence the backpropagation update is identical to the Counter-Hebb learning update, and we got an equivalence in the guided processing framework.

\subsection{The model heads}
\begin{figure}[ht]
  \centering
  \includegraphics[width=0.6\linewidth]{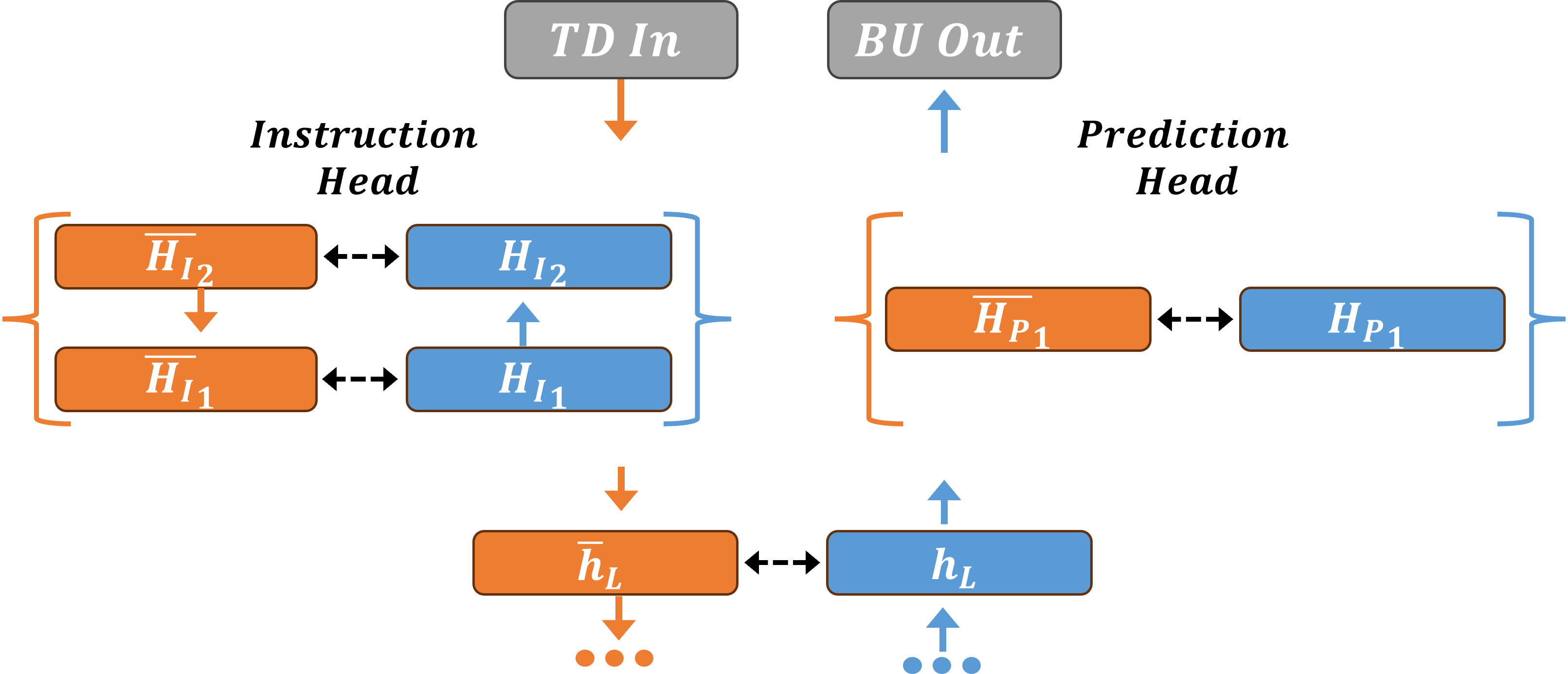}
  \caption{The figure depicts the prediction and instruction heads of the BU-TD model. Each head consists of two parts: one for the BU network and the other for the TD network. These parts maintain the symmetric structure with lateral connectivity of the BU-TD model. The instruction head employs a 2-layer MLP, while the prediction head utilizes a single linear layer. Only one head can be active in each pass of the BU-TD model, enabling selection between the instruction head and the prediction head. These heads can be alternated, with a different head chosen in each pass. The prediction head is responsible for model predictions. In the BU stream, it generates predictions based on input data, while in the TD stream, it delivers prediction error information. On the other hand, the instruction head bridges the instructional space with visual concepts. The TD stream maps task representations into the model's hidden space, while the BU stream maps the visual space into the instructional space. Refer to Fig \ref{fig: mtl algorithm} for an illustration of how the two heads are utilized in learning instruction-based models.}
  \label{fig: heads}
\end{figure}

In this section, we describe the heads structure of our proposed BU-TD model. The BU-TD core network consists of two symmetric neural networks that are connected via lateral connections, as described in ~\ref{section: the BU-TD model}. This core network is extended by two heads: a prediction head, and an instruction head, each employing an additional small BU-TD neural network, typically one to two-layer Multi-Layer Perceptron. Similar to the core, the heads consist of two connected parts: one for the BU network and the other for the TD network,  thus preserving the symmetrical structure and lateral connectivity of the BU-TD model. 
This results in two pairs of symmetric heads. The first pair is for the predictions: $H_{pred}$ in the BU stream, and its symmetric counterpart in the TD stream $\bar{H}_{pred}$. The second pair is for the instructions: $H_{instruct}$ in the BU stream, and its symmetric counterpart in the TD stream $\bar{H}_{instruct}$. Only one head can participate in each pass of the network (either BU pass or TD pass, for instance in the Counter-Hebb guided learning algorithm, the first TD pass uses the instruction head, while the following two passes (BU followed by a TD) use the prediction head. See Fig ~\ref{fig: heads} for an illustration of the heads' structure.

\subsection{Datasets}
\label{appendix: datasets}

Similar to other biologically motivated learning methods, we compare CH learning with BP on standard image classification benchmarks: MNIST \cite{lecun1998gradient}, Fashion-MNIST \cite{xiao2017fashion}, and CIFAR10 \cite{krizhevsky2009learning}. In addition, we use two common multi-task learning (MTL) benchmarks, the \textit{Multi-MNIST} \cite{sabour2017dynamic} dataset, and the \textit{CelebA} \cite{liu2015deep} dataset, to evaluate the ability of our model to guide the visual process according to instruction signals.

\textbf{Multi-MNIST}, introduced by \cite{sabour2017dynamic} and modified by \citet{sener2018multi}, is a simple two-task supervised learning benchmark dataset constructed by uniformly sampling two overlayed MNIST \cite{lecun1998gradient} digits. One digit is placed in the top-left corner, while the other is in the bottom-right corner. Each of the two overlaid images corresponds to a 10-class classification task. We generated the dataset using the code provided by \citet{kurin2022defense}, which samples the training set from the first 50,000 MNIST training images, and the test set from the original MNIST test set. We omitted the validation set, and the hyper-parameters were tuned based solely on the training set. 

The \textbf{CelebA} dataset \cite{liu2015deep} (with standard training, and test splits) comprises more than 200,000 face images of celebrities along with annotations for 40 attributes, such as the presence of eyeglasses, gender, smiling, and more. Within the context of Multi-Task Learning research, it is frequently approached as a 40-task classification challenge, where each task involves binary classification for one of the attributes.

\subsection{Experimental settings and results}
\label{appendix: exp}

\subsubsection{Computational resources}
All the experiments were conducted using either NVIDIA RTX 6000 GPU or NVIDIA RTX 8000 GPU. For all experiments but CelebA, a single NVIDIA RTX 6000 GPU was used, with the experiments utilizing only a fraction of its capacity. In the case of the CelebA dataset, either a single NVIDIA RTX 8000 GPU or two NVIDIA RTX 6000 GPUs were used.

\subsubsection{Image classification}
\label{appendix: stl}

We evaluated the Counter-Hebb learning, in the unguided regime, on the task of image classification, compared with backpropagation and other biologically plausible learning algorithms under the same settings. The following results extend the results shown in Table ~\ref{table: STL results} by evaluating the results obtained when learning with the MSE loss, in addition to the Cross-entropy loss reported in the main text. Note that there are two variations of CIM \cite{bozkurt2024correlative}, we report here the highest score obtained among the CIM experiments.

We repeat the same experiments as conducted in \citet{bozkurt2024correlative}, our BU network employs a two-layer fully connected network, with a hidden layer of size 500 for both MNIST and Fashion-MNIST datasets and size 1,000 for CIFAR10. The standard Adam optimizer \cite{ruder2016overview} was used to optimize both the Cross-Entropy loss and MSE loss without any regularization. We trained for $50$ epochs with an exponential learning rate decay with $\gamma=0.95$. The initial learning rate was $10^{-4}$, and the batch size $20$. All hyper-parameters but the initial learning rate were taken from the baseline experiments and were not optimized. The initial lr was selected from  $1 \cdot 10^{-3}, 5 \cdot 10^{-4}, 1 \cdot 10^{-4}$ according to the best test results on the CIFAR dataset   The results are presented in Table ~\ref{app table: STL results}.

\begin{table}[t]
    \caption{Unguided learning results: mean and standard deviation of the test accuracy (in percentages) across 10 runs. The proposed CH learning algorithm is compared with BP and other biological state-of-the-art methods. The baseline results were taken from \citet{bozkurt2024correlative}.}
    \label{app table: STL results}
    \begin{center}
    \begin{tabular}{llll}
        \toprule
        Method     & MNIST  & Fashion MNIST     & CIFAR10 \\
        \midrule
        CIM [\citeyear{bozkurt2024correlative}] & 97.71 $\pm$ 0.1 & 88.14 $\pm$ 0.3 & 51.86 $\pm$ 0.3 \\
        EP [\citeyear{scellier2017equilibrium}] & 97.61 $\pm$ 0.1 & 88.06 $\pm$ 0.7 & 49.28 $\pm$ 0.5 \\
        CSM [\citeyear{qin2021contrastive}] & 98.08 $\pm$ 0.1 & 88.73 $\pm$ 0.2 & 40.79 \\
        PC [\citeyear{whittington2017approximation}] & 98.17 $\pm$ 0.2 & 89.31 $\pm$ 0.4 & - \\
        PC-Nudge [\citeyear{millidge2022backpropagation}] & 97.71 $\pm$ 0.1 & 88.49 $\pm$ 0.3 & 48.58 $\pm$ 0.7 \\
        FA [\citeyear{lillicrap2016random}] (Cross-Entropy loss) & 97.95 $\pm$ 0.08 & 88.38 $\pm$ 0.9 & 52.37 $\pm$ 0.4 \\
        FA [\citeyear{lillicrap2016random}] (MSE loss) & 97.99 $\pm$ 0.03 & 88.72 $\pm$ 0.5 & 50.75 $\pm$ 0.4 \\
        BP (Cross-Entropy loss) & 98.27 $\pm$ 0.03 & 89.41 $\pm$ 0.2 & 53.96 $\pm$ 0.3 \\
        BP (MSE loss) & 97.58 $\pm$ 0.01 & 88.39 $\pm$ 0.1 & 52.75 $\pm$ 0.1 \\
        \midrule
        BP (ours) (Cross-Entropy loss) & 98.33 $\pm$ 0.04 & 89.94 $\pm$ 0.2 & 55.47 $\pm$ 0.3 \\
        CH Sym Init (Cross-Entropy loss) & 98.34 $\pm$ 0.06 & 89.99 $\pm$ 0.2 & 55.54 $\pm$ 0.3 \\
        CH Asym Init (Cross-Entropy loss) & 98.17 $\pm$ 0.06 & 89.27 $\pm$ 0.1 & 54.28 $\pm$ 0.2 \\
        \midrule
        BP (ours) (MSE loss) & 98.36 $\pm$ 0.08 & 90.16 $\pm$ 0.2 & 54.50 $\pm$ 0.4 \\
        CH Sym Init (MSE loss) & 98.37 $\pm$ 0.07 & 90.13 $\pm$ 0.2 & 54.56 $\pm$ 0.3 \\
        CH Asym Init (MSE loss) & 98.21 $\pm$ 0.06 & 89.54 $\pm$ 0.2 & 53.09 $\pm$ 0.3 \\
        \bottomrule
      \end{tabular}
    \end{center}
\end{table}

The results, shown in table ~\ref{app table: STL results}, empirically validate that CH learning is equivalent to BP in the symmetric case, and approximates BP in the asymmetric case for both Cross-Entropy and MSE loss. Furthermore, the proposed asymmetric CH learning method shows a significantly smaller gap from BP compared with the other biological learning methods.

\subsubsection{Image classification with convolutional networks}
\label{appendix: stl conv}

The above experiments show that asymmetric CH learning approximates backpropagation well for fully connected networks (Multi-Layer Perceptron). Since biologically plausible learning methods often struggle to scale to larger networks and other types of architectures such as convolutional networks \cite{ernoult2022towards}, we conducted additional experiments to assess the robustness of CH in other settings.  

For these experiments, we used the exact architecture and most hyper-parameters that were chosen for the Multi-MNIST benchmark. We ran two-layer convolutional networks on the MNIST and CIFAR10 benchmarks. Unlike the guided experiments, conducted on the Multi-MNIST benchmark, in these experiments, we used $32$ channels for the convolution layers and increased the batch size to $256$ for both MNIST and CIFAR10. The models were trained for $100$ epochs, although most converged much faster. 

To further examine the effect of symmetric weights, we propose a weak symmetry scenario, where the BU and TD weights are initialized symmetrically, but we introduce noise to the Counter-Hebb update, simulating a more realistic case of noisy update where the BU and TD weight adjustments are not identical. Hence, the weights do not maintain symmetry during the learning. It is worth noting that both this weak symmetric scenario and the asymmetric scenario, in which weights are initialized asymmetrically with close to zero values, lead to a near symmetry between the BU and TD weights after updates are applied. Therefore, the weak symmetric scenario can also approximate the following scenario where: 1) the weights are initialized asymmetrically with noise in the updates, and 2) the brain has already undergone some learning and synaptic updates before learning the new task, so the current state of the weights is not as random as in their initialization. In this setting, at each update step, the update value of each weight is multiplied by a random, relatively large, noise from a $\mathcal{N}(1,\,\sigma)$ distribution (a different random variable for each weight).

 In the below experiments, we keep all the settings above, and each time focus on a single parameter that is evaluated. The following parameters were examined:
\begin{itemize}
    \item the number of channels in each convolution layer
    \item a weight decay regularization term
    \item weak symmetric weights with different magnitudes of noise 
\end{itemize}

\begin{figure}[H]
    \centering
    \includegraphics[width=0.9\columnwidth]{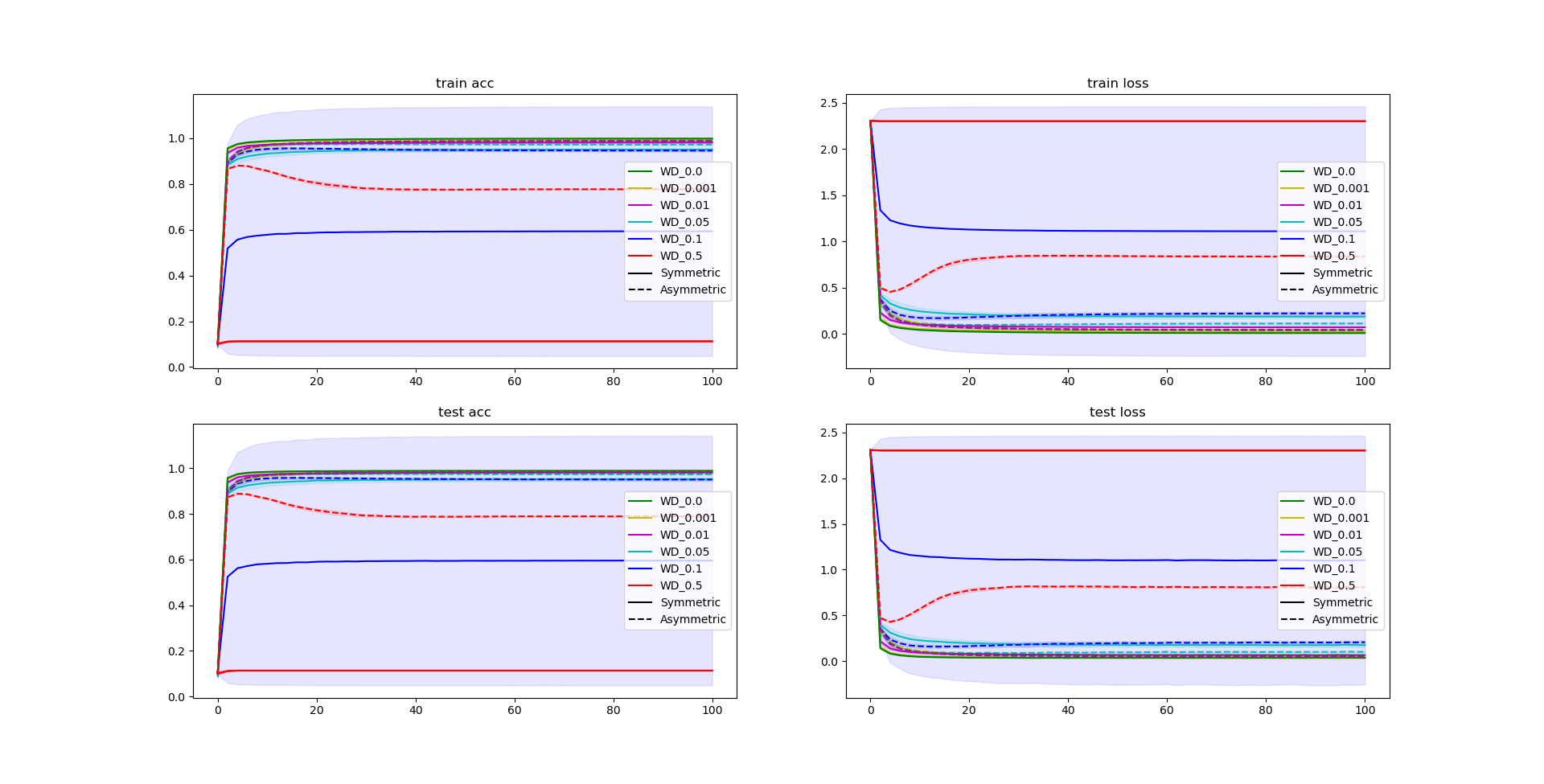}
    \caption{MNIST results: comparing different weight decay values and presenting the mean performance including std per training epoch averaged across 5 runs.}
    \label{results fig: mnist wd}
\end{figure}

\begin{figure}[H]
    \centering
    \includegraphics[width=0.9\columnwidth]{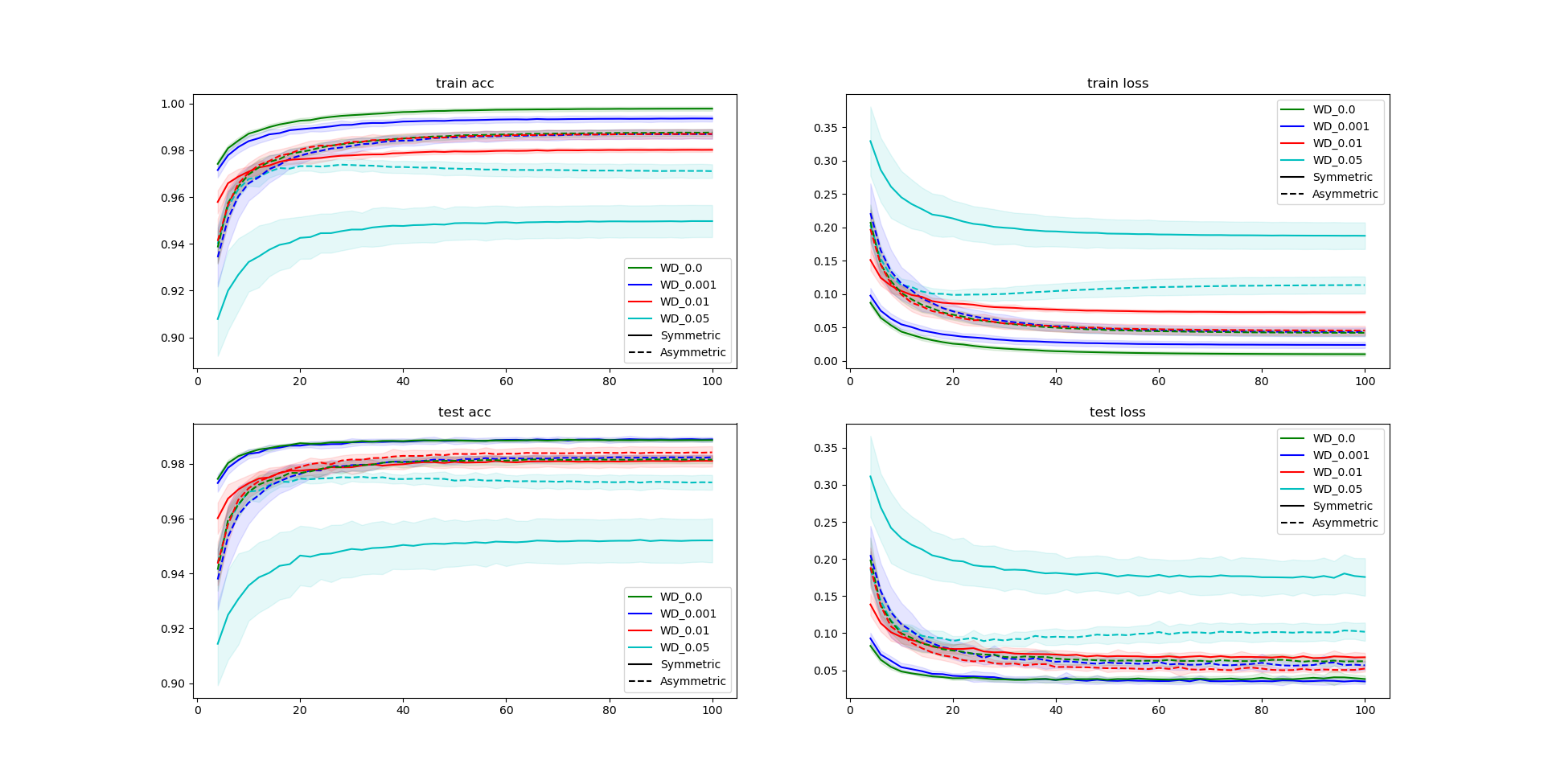}
    \caption{MNIST results: comparing different weight decay values and presenting the mean performance including std per training epoch averaged across 5 runs. Focusing on less weight decay factors, and starting from the 4th iteration for better visualization of the differences}
    \label{results fig: mnist wd less}
\end{figure}

\begin{figure}[H]
    \centering
    \includegraphics[width=0.9\columnwidth]{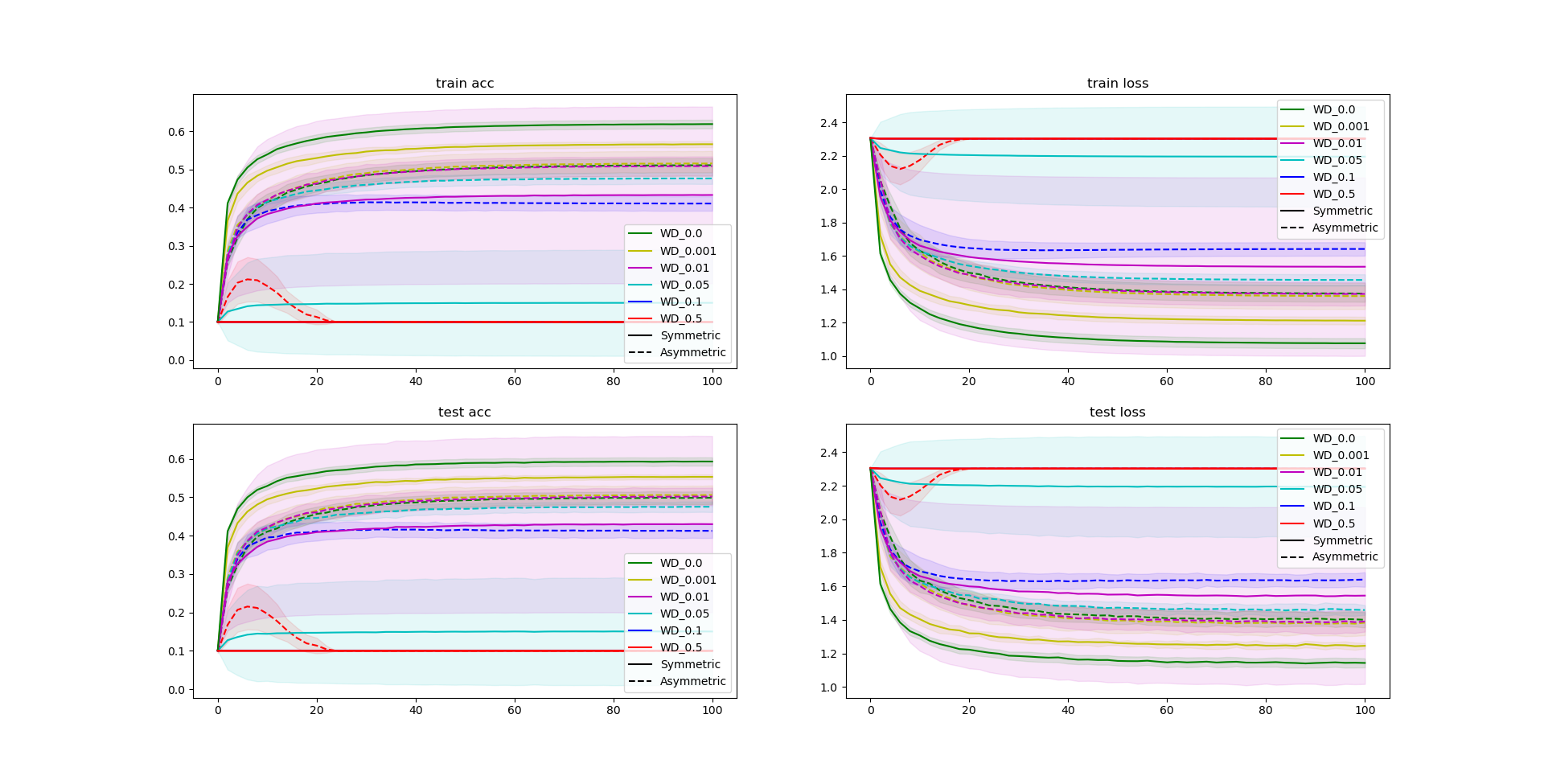}
    \caption{CIFAR10 results: comparing different weight decay values and presenting the mean performance including std per training epoch averaged across 5 runs.}
    \label{results fig: cifar wd}
\end{figure}

\begin{figure}[H]
    \centering
    \includegraphics[width=0.9\columnwidth]{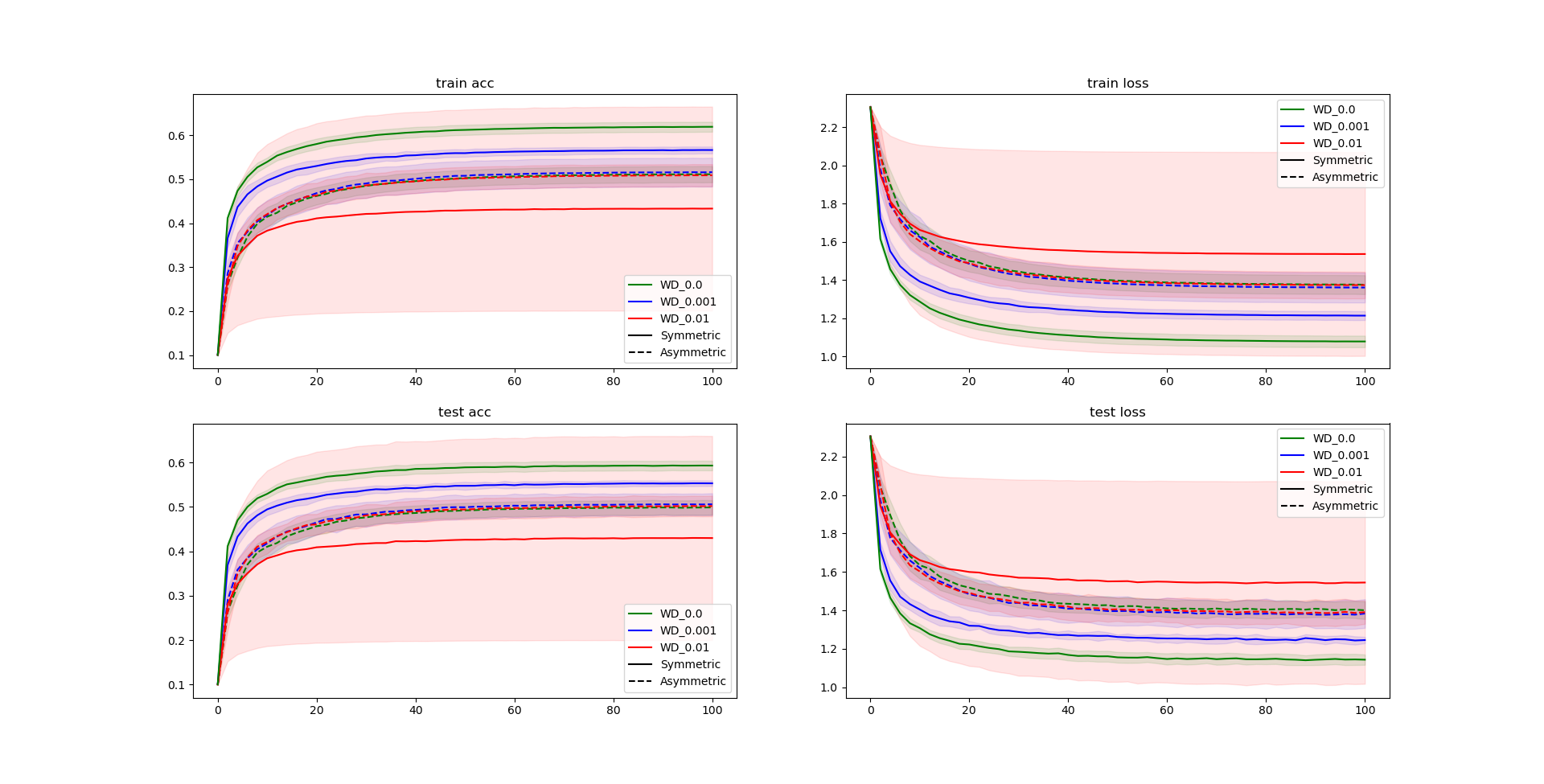}
    \caption{CIFAR10 results: comparing different weight decay values and presenting the mean performance including std per training epoch averaged across 5 runs. Focusing on less weight decay factors for better visualization of the differences}
    \label{results fig: cifar wd less}
\end{figure}

The results shown in Figures ~\ref{results fig: mnist wd}, ~\ref{results fig: mnist wd less}, ~\ref{results fig: cifar wd}, ~\ref{results fig: cifar wd less}, compare different weight decay values and show that CH learning in the asymmetric case approximates the symmetric case (which is equivalent to backpropagation). Furthermore, in the backpropagation case, we observed a significant degradation in performance when increasing the weight decay term, up to the level where the model is not learning and the performances are near chance. Surprisingly, the asymmetric case is much more robust to this effect. 

\begin{figure}[H]
    \centering
    \includegraphics[width=0.9\columnwidth]{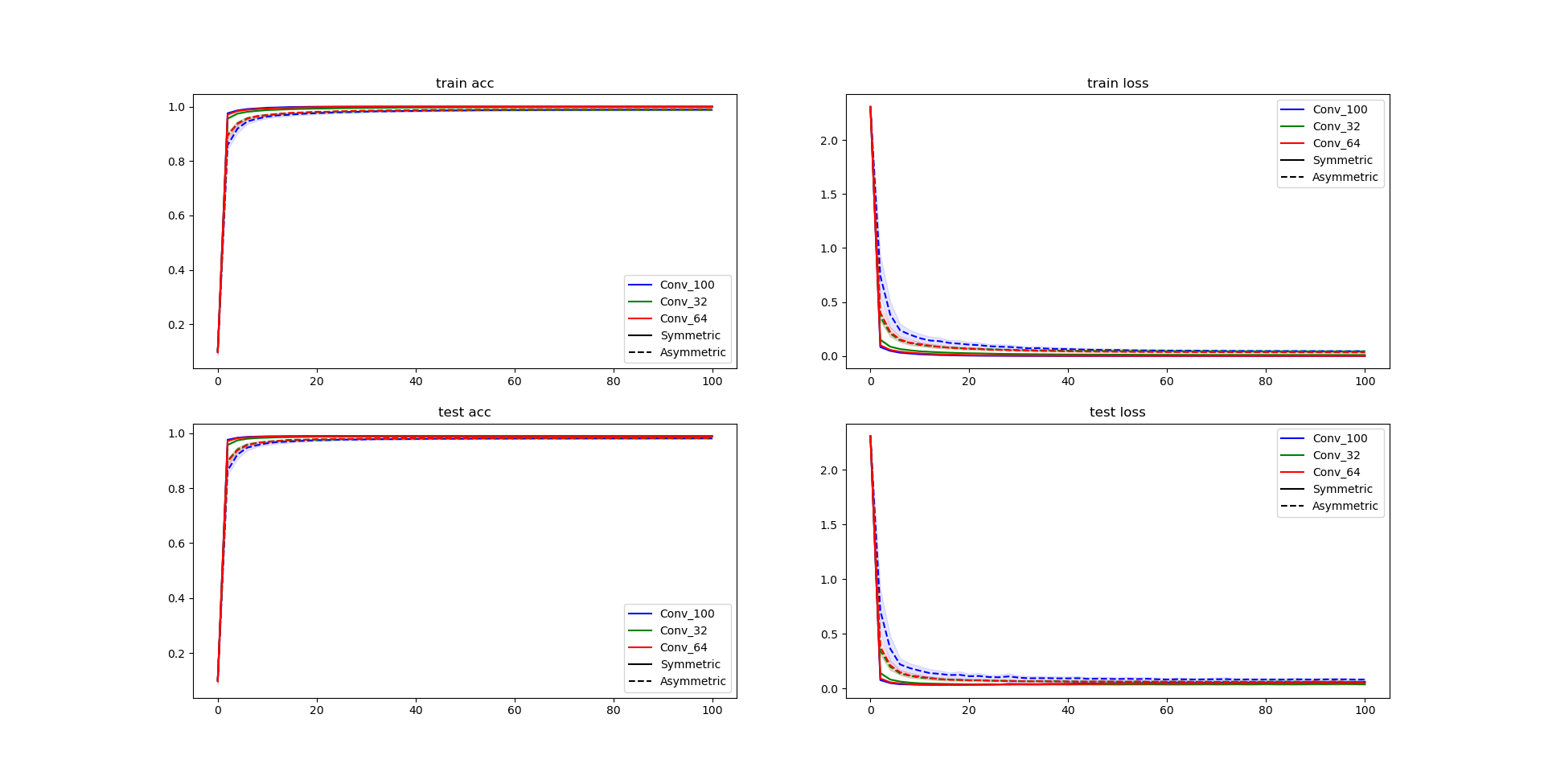}
    \caption{MNIST results: comparing different channels and presenting the mean performance including std per training epoch averaged across 5 runs.}
    \label{results fig: mnist channels}
\end{figure}

\begin{figure}[H]
    \centering
    \includegraphics[width=0.9\columnwidth]{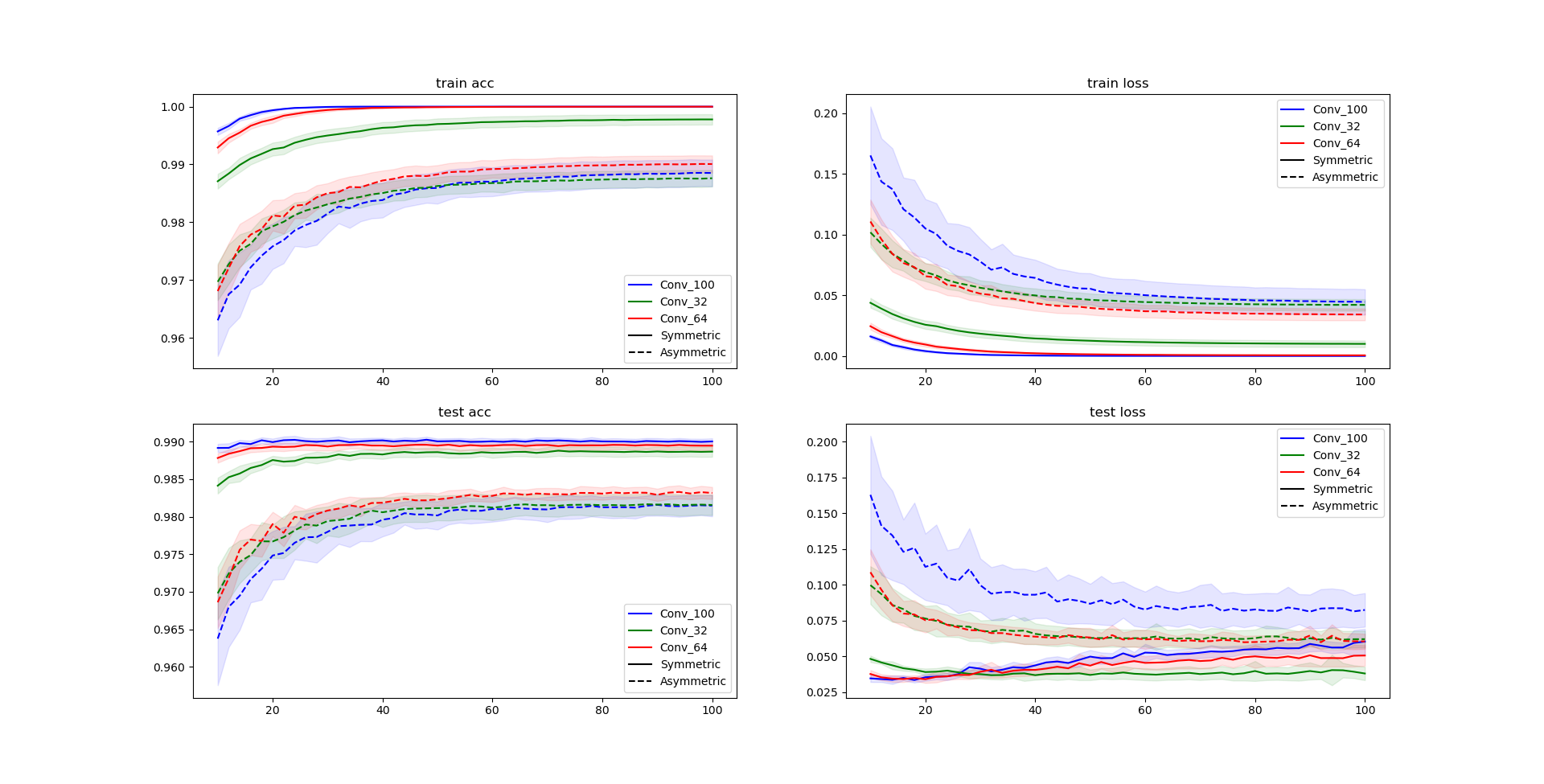}
    \caption{MNIST results: comparing different channels and presenting the mean performance including std per training epoch averaged across 5 runs. Starting from the 10th iteration for better visualization of the differences}
    \label{results fig: mnist channels start 10}
\end{figure}

\begin{figure}[H]
    \centering
    \includegraphics[width=0.9\columnwidth]{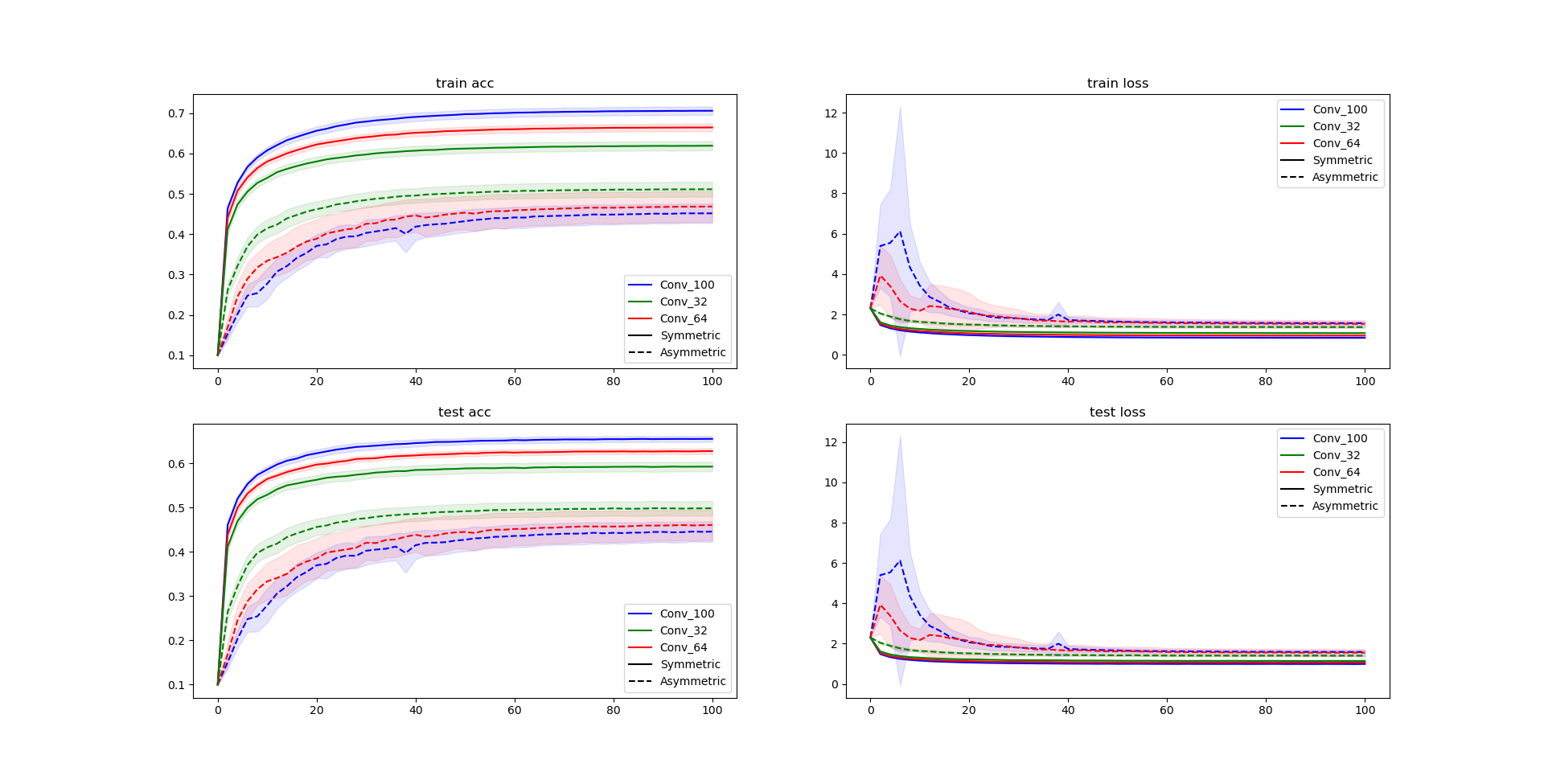}
    \caption{CIFAR results: comparing different channels and presenting the mean performance including std per training epoch across 5 runs.}
    \label{results fig: cifar channels}
\end{figure}

The results shown in Figures ~\ref{results fig: mnist channels}, ~\ref{results fig: mnist channels start 10}, ~\ref{results fig: cifar channels}, compare different numbers of channels for each convolution layer. Similar to prior works \cite{ernoult2022towards}, we observe that the gap between the symmetric and non-symmetric case increases as the capacity of the network increases, indicating that in large-scale tasks, backpropagation performs better than the asymmetric case.

\begin{figure}[H]
    \centering
    \includegraphics[width=0.9\columnwidth]{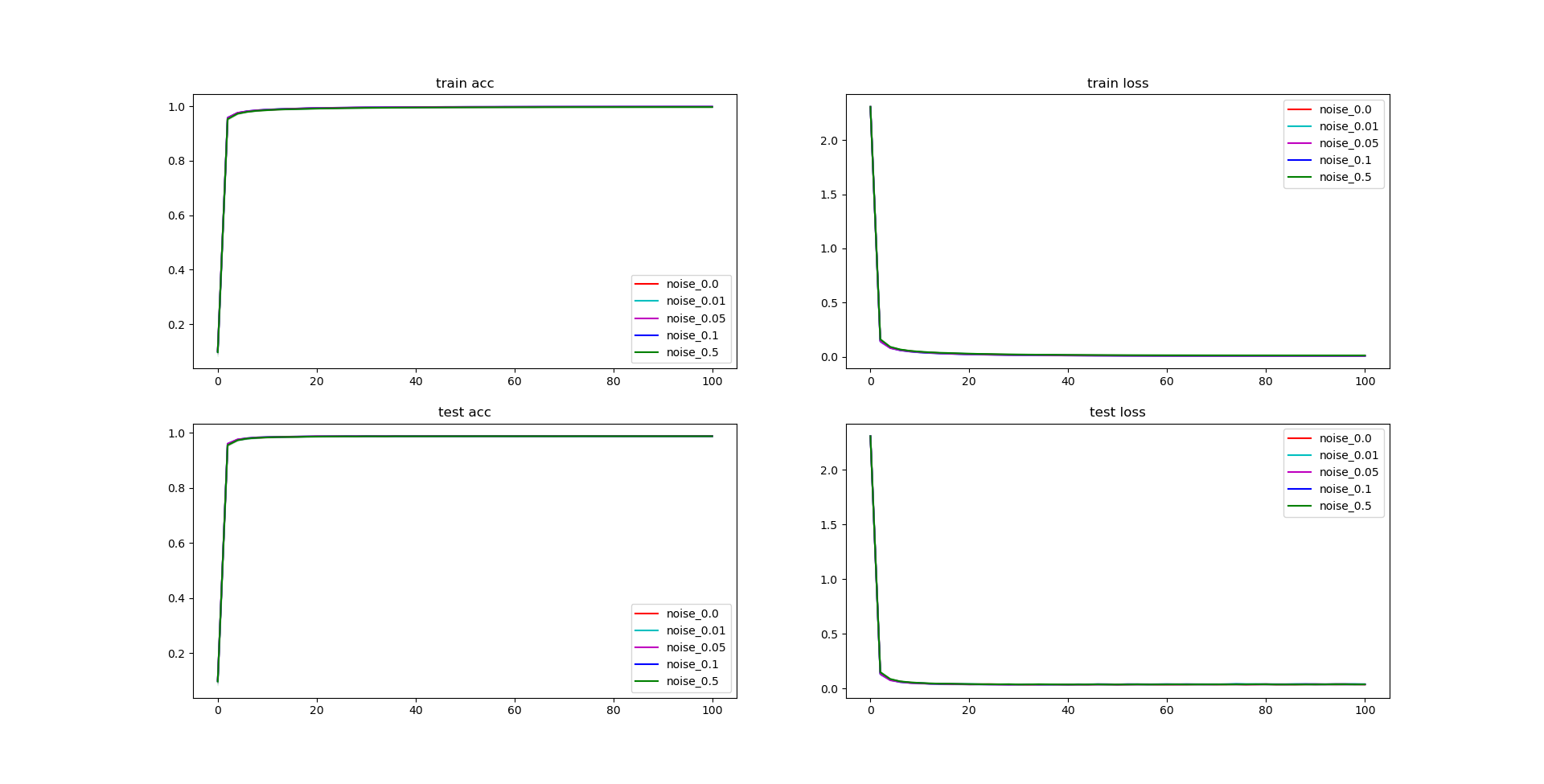}
    \caption{MNIST results: comparing different magnitudes of noise and presenting the mean performance including std per training epoch averaged across 5 runs.}
    \label{results fig: mnist weaksym}
\end{figure}

\begin{figure}[H]
    \centering
    \includegraphics[width=0.9\columnwidth]{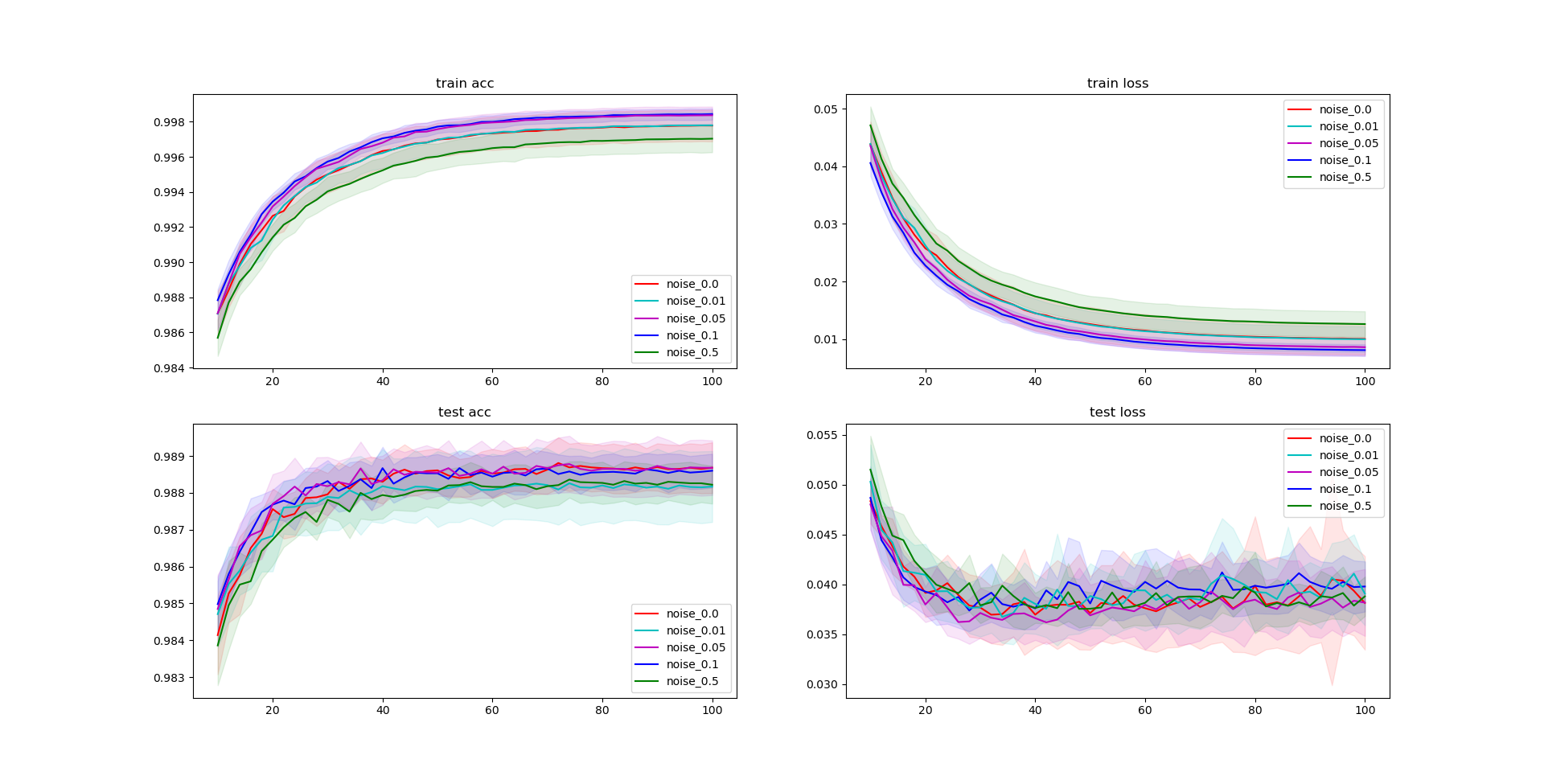}
    \caption{MNIST results: comparing different magnitudes of noise and presenting the mean performance including std per training epoch averaged across 5 runs. Starting from the 10th iteration for better visualization of the differences}
    \label{results fig: mnist weaksym start 10}
\end{figure}

\begin{figure}[H]
    \centering
    \includegraphics[width=0.9\columnwidth]{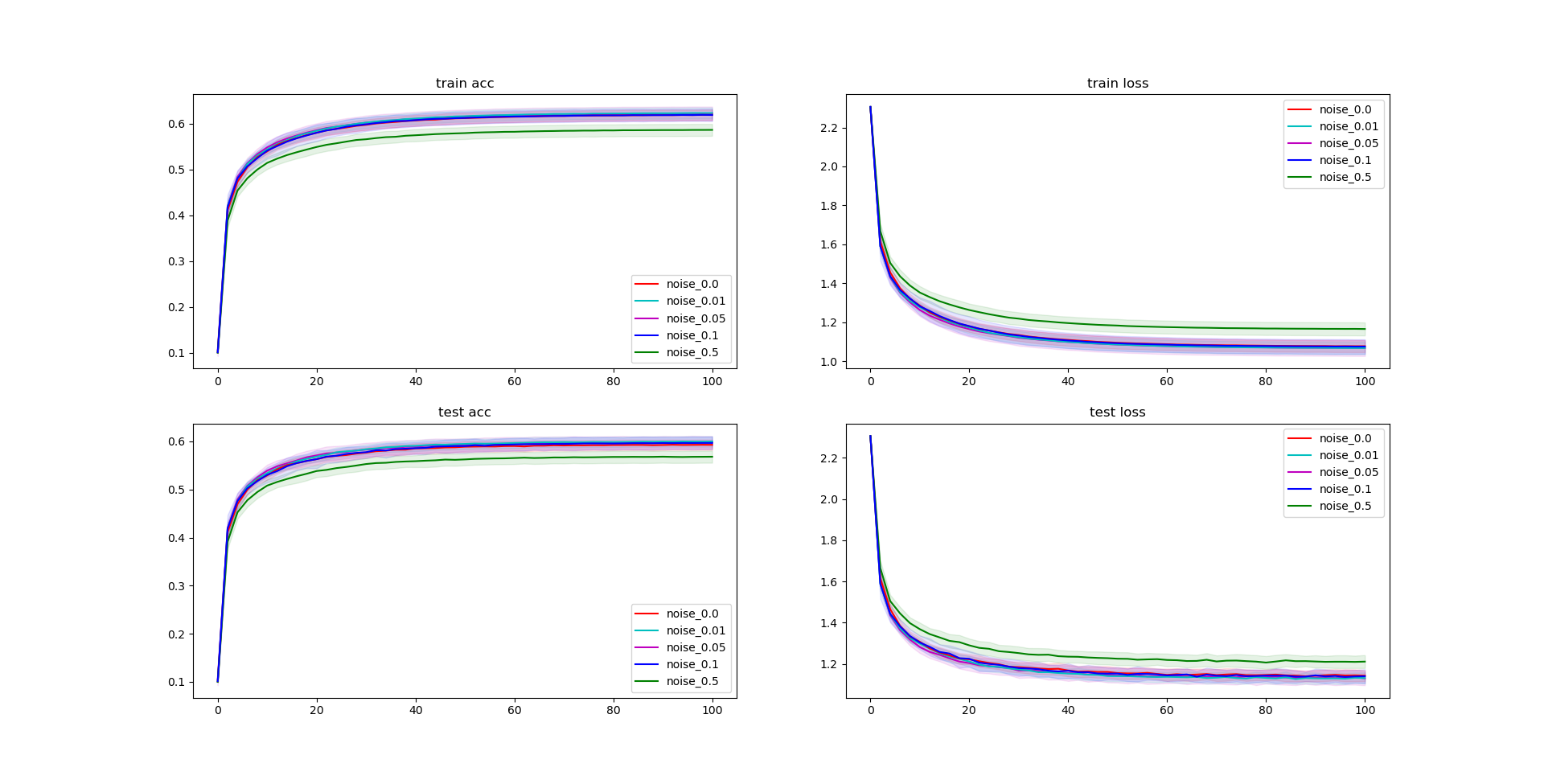}
    \caption{CIFAR results: comparing different magnitudes of noise and presenting the mean performance including std per training epoch averaged across 5 runs.}
    \label{results fig: cifar weaksym}
\end{figure}

\begin{figure}[H]
    \centering
    \includegraphics[width=0.9\columnwidth]{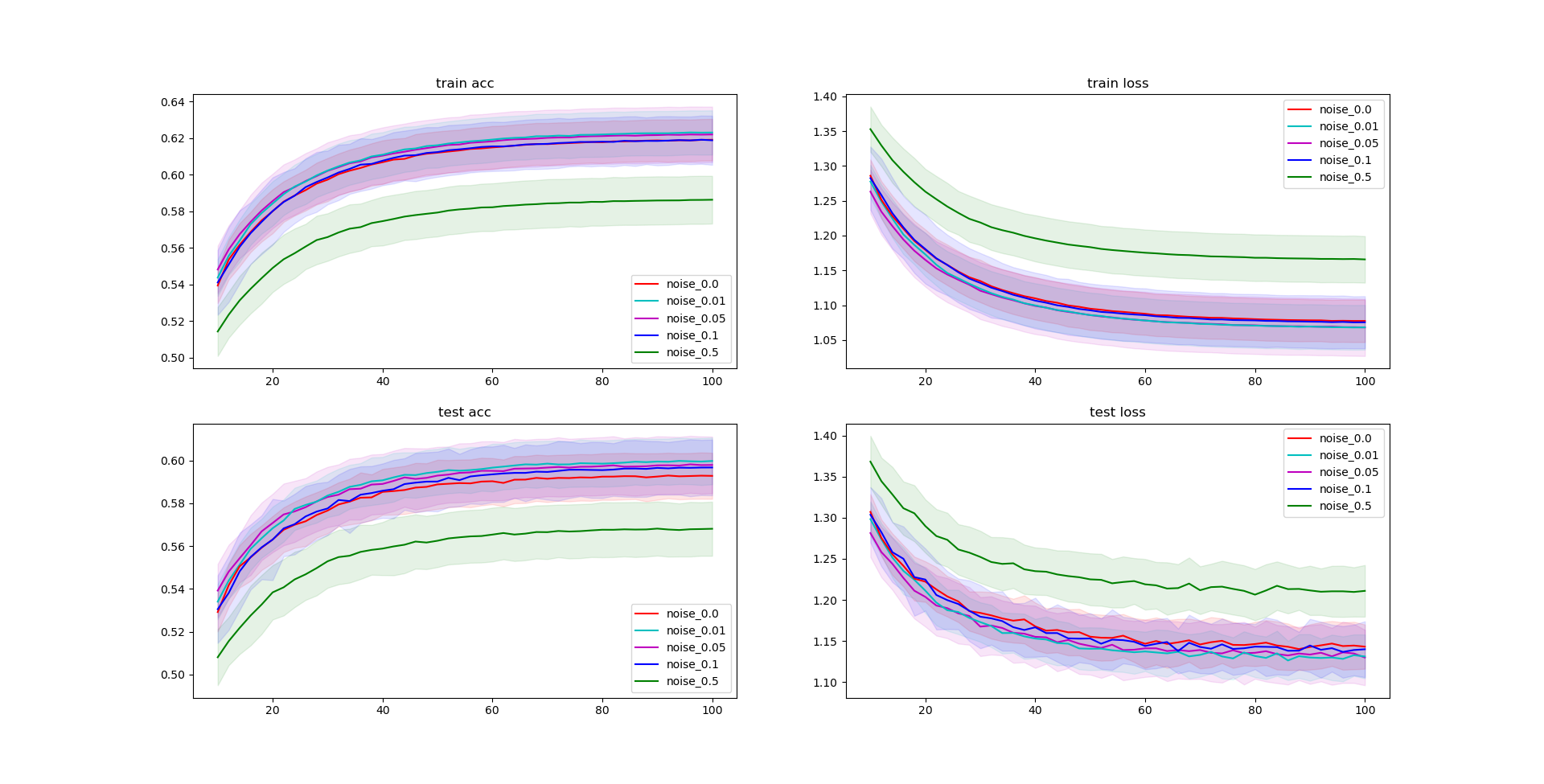}
    \caption{CIFAR results: comparing different magnitudes of noise and presenting the mean performance including std per training epoch averaged across 5 runs. Starting from the 10th iteration for better visualization of the differences}
    \label{results fig: cifar weaksym start 10}
\end{figure}

The results shown in Figures ~\ref{results fig: mnist weaksym}, ~\ref{results fig: mnist weaksym start 10}, ~\ref{results fig: cifar weaksym}, ~\ref{results fig: cifar weaksym start 10} compare different magnitudes of noise applied to the Counter-Hebb update. It is shown that a weak symmetry is sufficient for achieving a similar performance as backpropagation (referred to as a 0 noise in the figures), even with a relatively large magnitude of noise. Since as the training progresses the results gradually become less symmetric, it demonstrates the ability of CH learning to converge to good solutions even when the BU and TD weights are not symmetric and have different values. Thus, we believe that studying various weight initialization schemes could improve performance and close the gap between learning with symmetric and asymmetric weights.

\subsubsection{Guided visual processing settings}

In the guided experiments, we evaluate our model on Multi MNIST and CelebA, two common multi-task learning (MTL) benchmarks. Since current biological methods are not capable of guided processing, we compare CH with state-of-the-art non-biological optimization methods as reported by \citet{kurin2022defense}, replicating their setup and use their reported results. 

However, in contrast to the baseline multi-task learning methods, we do not use any learning 'tricks' such as dropout layers, regularization, or special optimizers. Instead, our model is trained straightforwardly using the Adam optimizer \cite{ruder2016overview}. Another distinction from the baseline methods is that we do not use a validation set. As a result, we do not use an early stopping mechanism, and report the final results obtained from the last epoch which might introduce some noisiness. Furthermore, all hyper-parameters were only lightly tuned based solely on the training set.

\subsubsection{Multi MNIST}
Similar to the baseline experiments conducted in \cite{kurin2022defense}, our BU network employs a simple architecture composed of $2$ convolutional layers followed by a single fully-connected layer and ReLU non-linearity, along with an additional fully-connected layer as the decoder. Each convolution layer includes $100$ channels, and a $5\times5$ kernel (a single stride and no padding). Similar to the baseline, the last fully connected layer size is $50$. Additionally, to support the BU-TD structure, we replace all max-pool layers with strided convolution layers. The strided convolution layers have $2\times2$ kernel size with a stride of $2$ (similarly to the max pool operation).

The standard Adam optimizer \cite{ruder2016overview} was used to optimize the Cross-Entropy loss without any regularization, as opposed to the baseline. Similar to the compared methods, we trained for $100$ epochs with an exponential learning rate decay with $\gamma=0.95$. The initial learning rate was $5 \cdot 10^{-4}$, and the batch size was $64$. We have chosen the learning rate from $[0.005, 0.001, 0.0005, 0.0001]$ based on the convergence rate on the train set.

 In the below experiments, we keep all the settings above, and each time focus on a single parameter that is evaluated. The following parameters were examined:
\begin{itemize}
    \item the number of channels in each convolution layer
    \item weak symmetric weights with different magnitudes of noise 
\end{itemize}

\begin{figure}[H]
    \centering
    \includegraphics[width=0.9\columnwidth]{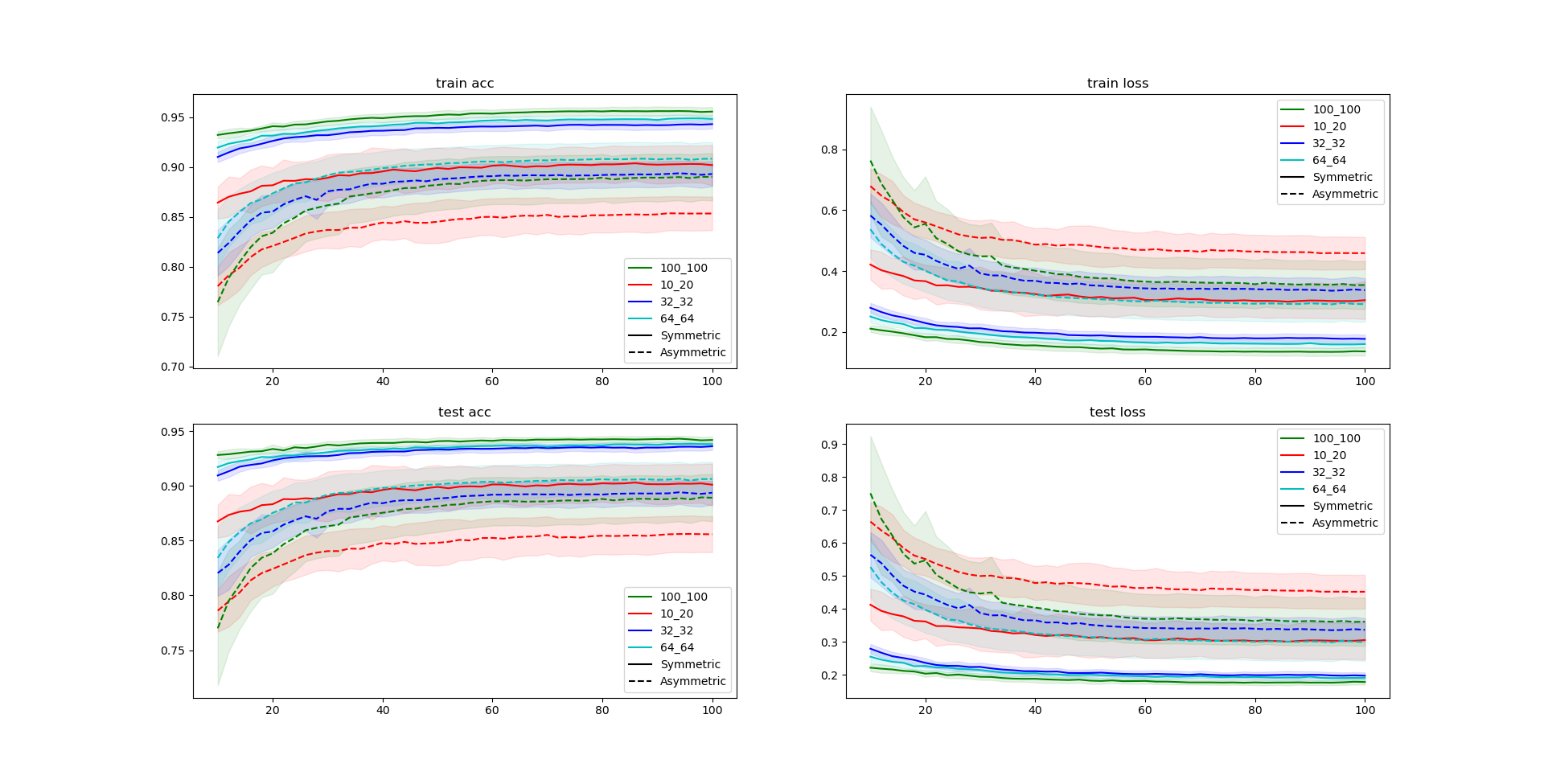}
    \caption{Multi-MNIST results: mean and 95\% confidence interval of the average task accuracy and loss per training epoch (starting from the 10th). Comparing different numbers of channels. }
    \label{results fig: multi mnist channels}
\end{figure}

\begin{figure}[H]
    \centering
    \includegraphics[width=0.9\columnwidth]{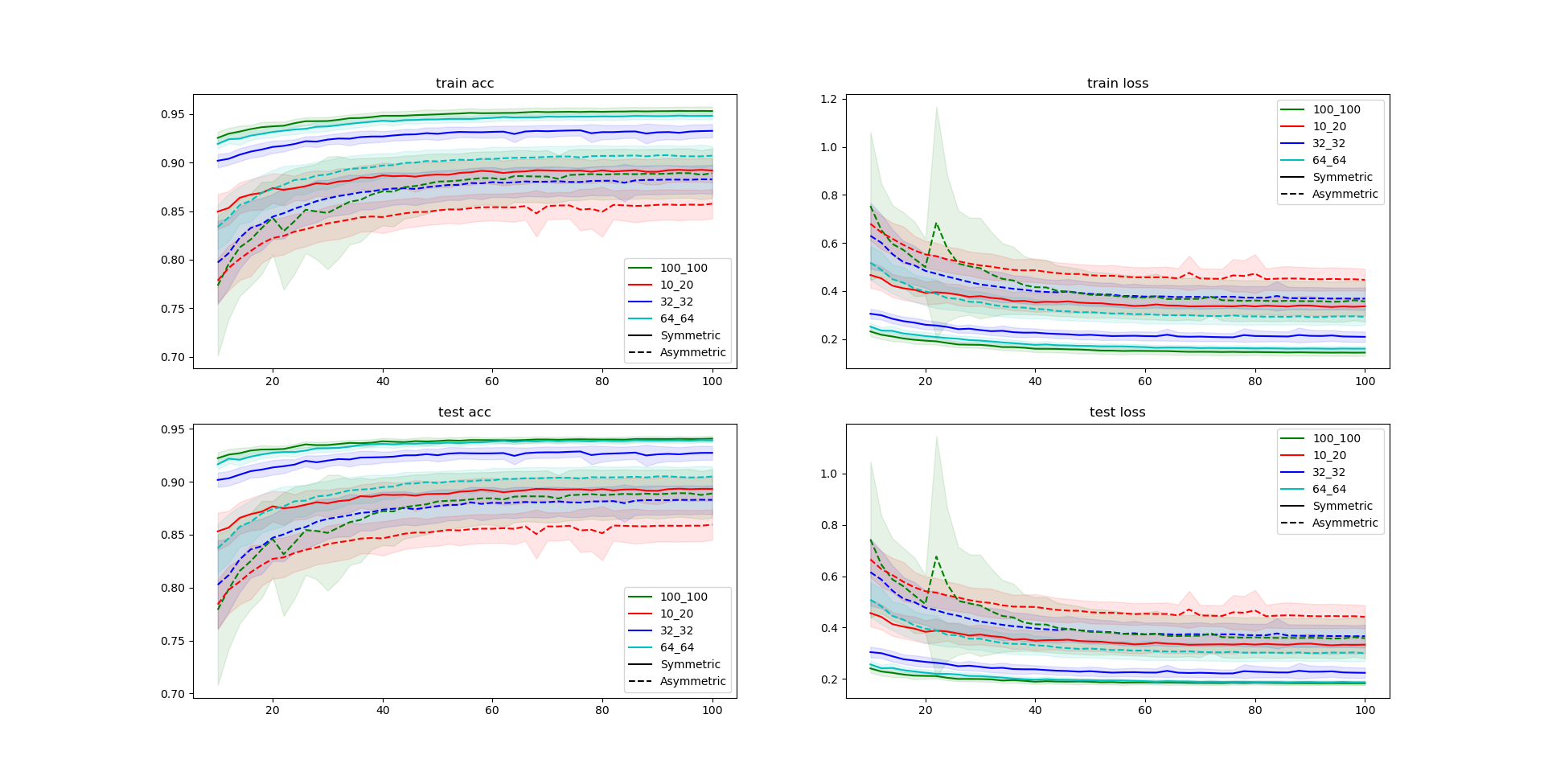}
    \caption{Multi-MNIST (single decoder) results: mean and 95\% confidence interval of the average task accuracy and loss per training epoch (starting from the 10th). Comparing different number of channels when the model uses a single decoder.}
    \label{results fig: multi mnist single decoder}
\end{figure}

\begin{figure}[H]
    \centering
    \includegraphics[width=0.9\columnwidth]{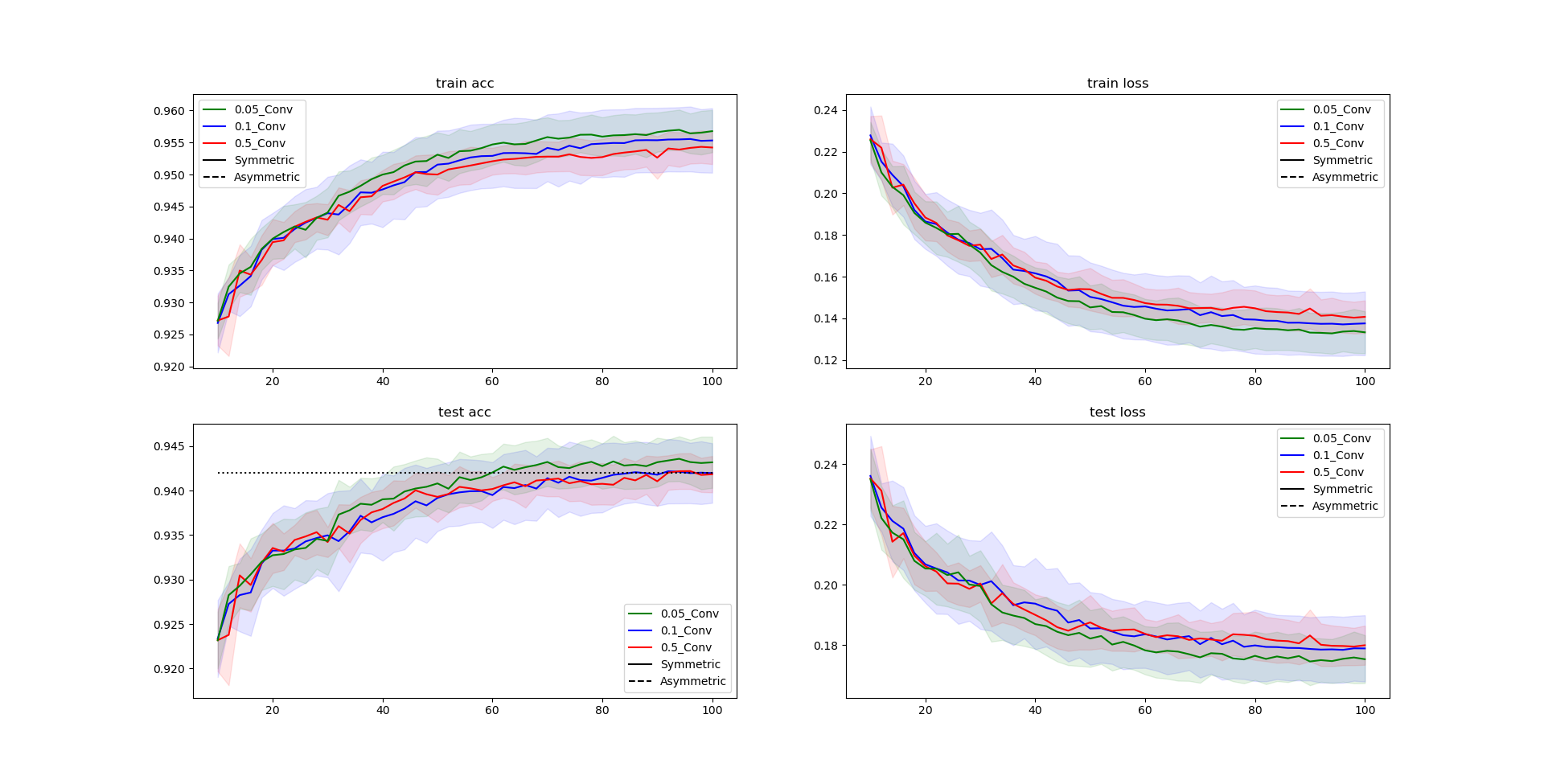}
    \caption{Multi-MNIST (weak symmetry) results: mean and 95\% confidence interval of the average task accuracy and loss per training epoch (starting from the 10th). Comparing different magnitudes of noise in the weak symmetry case.}
    \label{results fig: multi mnist weak sym}
\end{figure}

The results shown in Figures ~\ref{results fig: multi mnist channels}, ~\ref{results fig: multi mnist single decoder},  ~\ref{results fig: multi mnist weak sym}, demonstrate the robustness of the instruction-based method in a vanilla setting (as opposed to the baselines compared in the main text). 

We observe that increasing the capacity of the model (number of channels) (Fig ~\ref{results fig: multi mnist channels}) increases the performance. Interestingly, in the asymmetric case, further increasing the capacity beyond a certain point reduces the performance. 

Furthermore, when the model uses a single decoder (Fig ~\ref{results fig: multi mnist single decoder}) for both tasks, similar performances are maintained. This is a very important finding that highlights the ability of the TD stream to guide the BU stream. When employing a single decoder, the same network is used for both tasks (there are no task-specific parameters), thus the model cannot rely on task-specific parameters to handle multiple tasks. 

Moreover, the weak symmetry case (Fig ~\ref{results fig: multi mnist weak sym}) shows that an exact symmetry is not required by our model to perform similarly to backpropagation. The weak symmetry experiments achieve competitive (and even slightly higher) performances compared to exact symmetric weights. The dashed line indicates the final performance of the symmetric case.

\subsubsection{CelebA}

As done in previous work \cite{kurin2022defense}, we employ a ResNet-18 \cite{he2016deep} (without the final layer) with batch normalization layers \cite{ioffe2015batch}, and the decoder is a single linear fully-connected layer with a single neuron output for binary classification. Additionally, we remove the last average pooling layer to support the symmetric BU-TD structure. 

Batch normalization operates without reliance on learnable parameters, instead utilizing aggregated statistics such as the mean activation value of neurons across multiple iterations. Consequently, we implement distinct batch normalization for the BU and TD networks, with each network gathering statistics relevant to its own operations.

The standard Adam optimizer \cite{ruder2016overview} was used to optimize the Binary-Cross-Entropy loss without any regularization. Similar to the compared methods, we trained for $50$ epochs with an exponential learning rate decay with $\gamma=0.95$. The initial learning rate was $5 \cdot 10^{-4}$ which is chosen from $[0.005, 0.001, 0.0005, 0.0001]$ based on the convergence rate on the train set. The batch size is slightly smaller than the baselines in order to fit the GPU memory, and is set to $100$ when the BU and TD networks share the same set of weights, and $64$ otherwise. 

\begin{figure}[H]
    \centering
    \includegraphics[width=0.9\columnwidth]{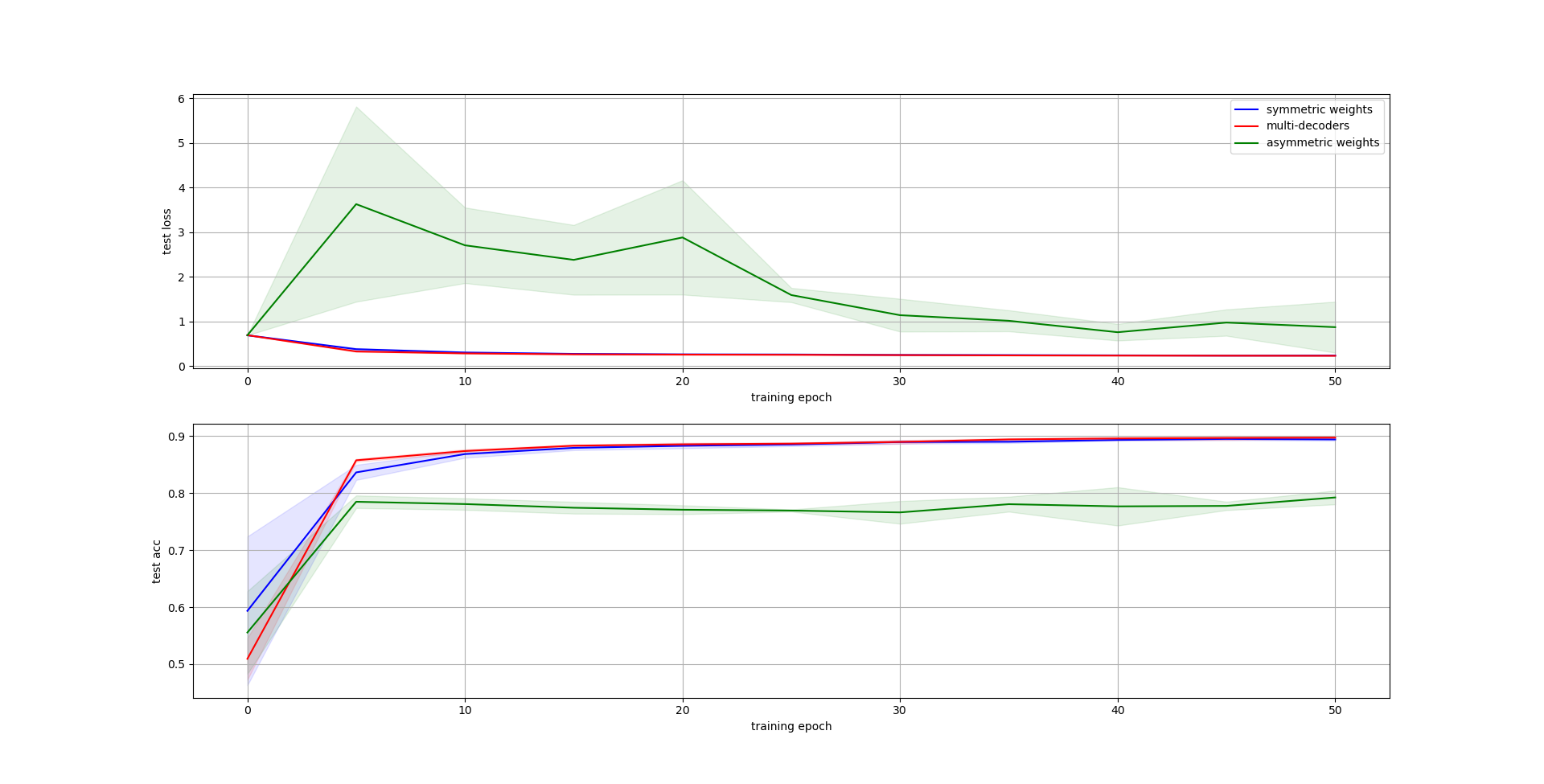}
    \caption{CelebA results: mean and std of the average task accuracy and loss on the test set per training epoch (sampled every 5 epochs).}
    \label{fig: celeba MTL results per epoch}
\end{figure}

\begin{figure}[H]
    \centering
    \includegraphics[width=0.9\columnwidth]{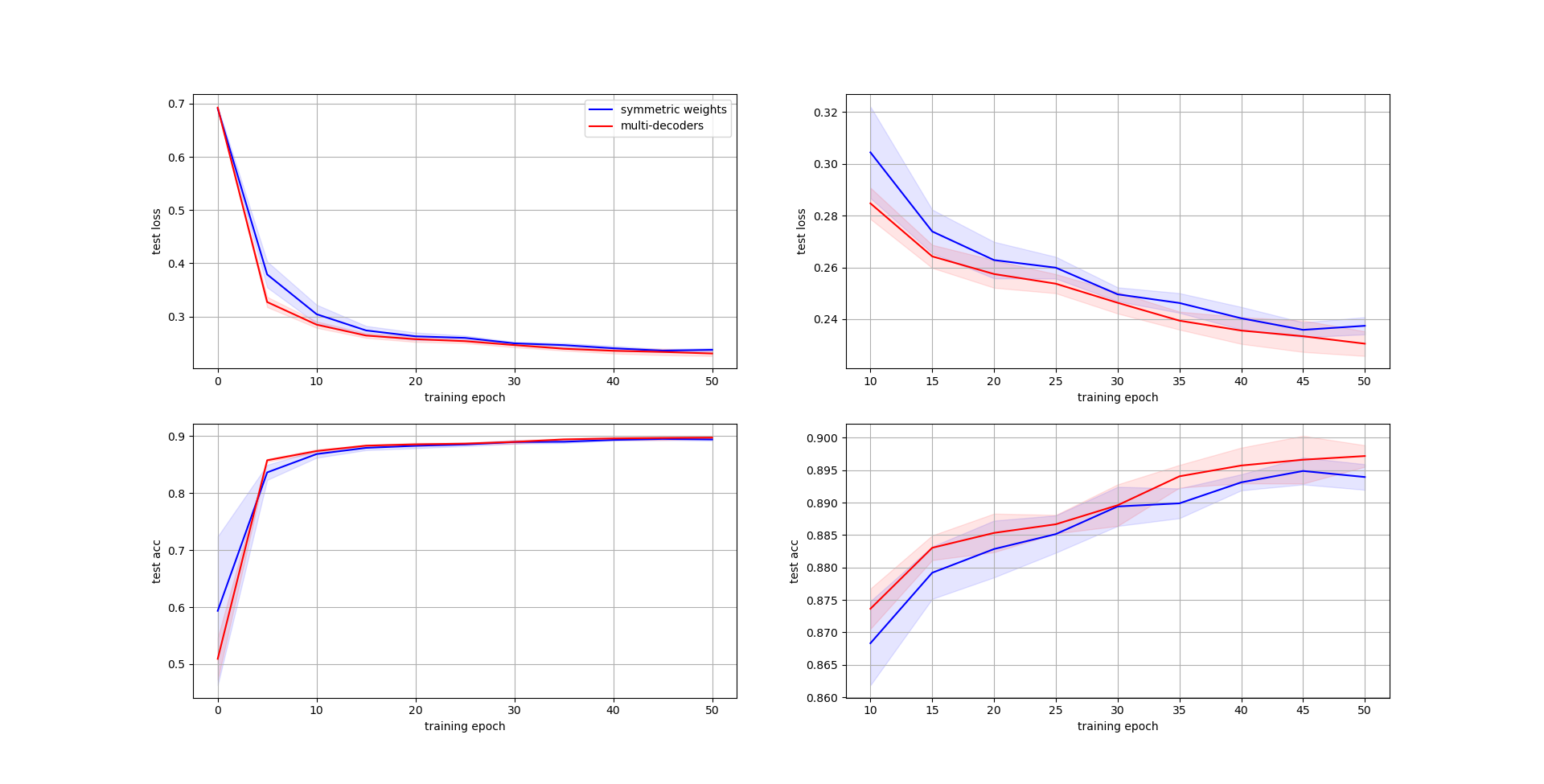}
    \caption{CelebA results: mean and std of the average task accuracy and loss on the test set per training epoch (sampled every 5 epochs). Within this figure, we omit the results of the asymmetric model due to being far from the other models. This allows for clearer observation of the distinctions among the remaining models. Additionally, on the right, presented the results starting from the 10th epoch.}
    \label{fig: celeba MTL results per epoch filter asym}
\end{figure}

Statistics of the average task test accuracy obtained from the last epoch (no early stopping) with $5$ repetitions are reported in table \ref{appendix: celeba results table}. We compare the BU-TD model across multiple configurations, as described in Appendix ~\ref{appendix: exp}. In addition, we plot the test results during the training process, sampled every 5 epochs, in Figures ~\ref{fig: celeba MTL results per epoch}, ~\ref{fig: celeba MTL results per epoch filter asym}.

\begin{table}[H]
\centering
\begin{tabular}{lll}
    \toprule
    Method     & CelebA Test Accuracy \\
    \midrule
    symmetric weights  & 89.51 $\pm$ 0.21  \\
    multi-decoders & 89.69 $\pm$ 0.12  \\
    asymmetric weights & 79.25 $\pm$ 1.63 \\ 
    \bottomrule
  \end{tabular}
\caption{CelebA results: mean and 95\% confidence interval of the avg. task test accuracy (in percentages) across 5 runs. See Appendix ~\ref{appendix: exp} for details}
\label{appendix: celeba results table}
\end{table}

The results on the CelebA dataset, which is more challenging are consistent with the Multi-MNIST results, demonstrating the BU-TD model's ability to scale and solve complex tasks through Counter-Hebb learning.

\subsection{Asymmetric Weights}
\label{appendix: asymmetric results}

Following the discussion in section ~\ref{section CH learning} Counter-Hebb scheme is guaranteed to converge to symmetric weights when applying weight decay, as the initial value of the weights gradually zeroed. However initial convergence of the model can require long training, beyond the training on a small number of tasks as in our experiments. Moreover, we demonstrate that attaining convergence to symmetric weights by itself does not suffice to yield favorable outcomes. This is evidenced by the model's inability to attain high performance, even when the weights approximate symmetry by the end of training. This mirrors the significance of initialization techniques in backpropagation in general.  

We hypothesize that the process of simultaneously learning numerous tasks, which is plausible in human learning, can lead to the convergence of "good" symmetric weights. The rationale is that in our case, the relevant sub-networks constitute only a portion of the model that is being used during the learning. Engaging in the learning of numerous tasks prompts the activation of the complete network model, with updates originating from diverse sources that can balance each other.

The convergence of the BU and TD to be symmetric, regardless of their initial state, enables simulating a mature network model by initialization of nearly symmetric weights from the beginning. This weak symmetry will be maintained through the symmetric update rule. We accomplish this by initializing with symmetric weights and incorporating noise into the update rule, leading to an approximation of symmetry from the second learning iteration onward. This scenario of weakly symmetric weights yields performances that are on par with the case of the symmetric weight, as shown in Figure ~\ref{results fig: multi mnist weak sym}.

\subsection{Sub-networks analysis}
\label{appendix: sub-networks analysis}

\begin{figure}[H]
    \centering
    \includegraphics[width=0.9\columnwidth]{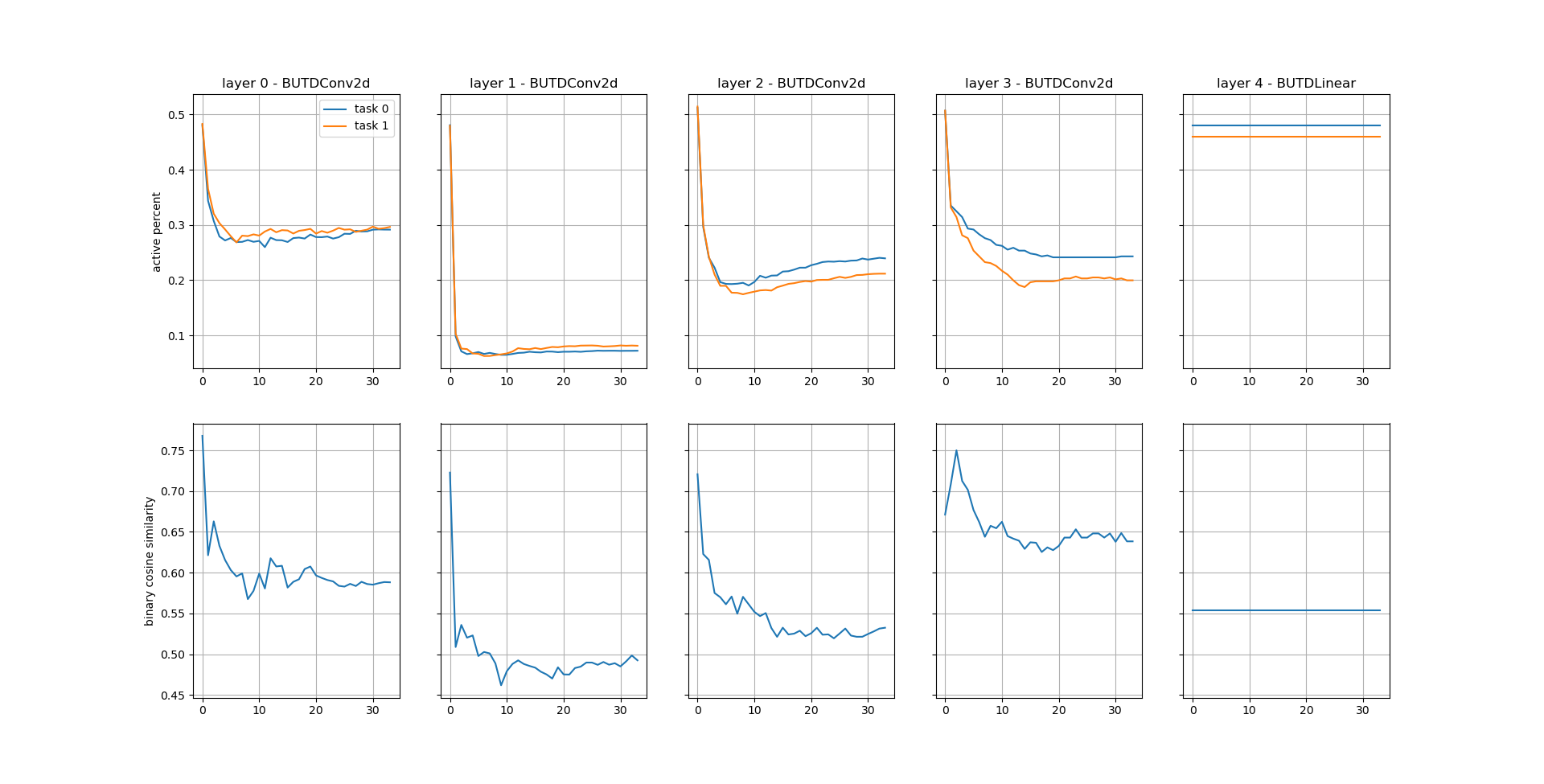}
    \caption{Analysis of the different sub-networks learned by the model and how they evolved during the learning process. We sampled the task-dependent sub-networks for each task Every 3 epochs during learning the Multi-MNIST data set. The columns in the figures represent the different hidden layers in the network, ordered from the first hidden layer on the left to the top hidden layer on the right. The X-axis of all figures represents the epochs during training. The first row shows the percentage of neurons that are being used in each sub-network for every layer. The second row shows the cosine similarity between the binary masking vectors of the two tasks, where 1 indicates an active neuron that is being used in the sub-network and 0 denotes an inactive one.}
    \label{fig: sub-networks analysis}
\end{figure}

In our experiments, the model has exhibited the capability to solve multiple tasks by assigning a distinct task-specific sub-network for each task. In this section, we analyze the resulting sub-networks focusing on the Multi-MNIST experiment, where two tasks: "left" and "right" are involved. For the purpose of this analysis, we have evaluated a BU-TD model with symmetric weights and a single decoder. Our analysis shows the characteristics of the different sub-networks learned by the model and how they evolved during the learning process. Specifically, we have extracted for each task its corresponding sub-network every 3 epochs. Then we evaluated the size of the sub-networks and examined the level of overlap between them. The analysis is presented in Fig ~\ref{fig: sub-networks analysis}. The findings collectively provide insights into the learning dynamics of the model and its ability to develop task-specific representations. 

From the figure, several findings can be drawn:
\begin{itemize}
    \item Dynamic Nature of Sub-Networks: The sub-networks exhibit changes throughout the learning process, indicating that the model adapts and refines its sub-networks representations. This adaptation occurs especially in the earlier epochs of the training.
    \item Sparsity in Sub-Networks: A notable characteristic of the sub-networks is their sparsity (row 1)- a small percentage of active neurons. The percentage of active neurons drastically decreases at the early iterations until reaching a plateau. The level of sparsity is lower at the first layer as it represents the image signal and needs to capture a large number of pixels. 
    \item Fixed top-level hidden layer: The top-level hidden layer is obtained by passing the task via the task head function. Since we do not update the task head during the training, This layer remains fixed during the learning. 
    \item Similarity Between Sub-Networks: Despite the sparsity of the sub-networks, they demonstrate some degree of similarity (row 2). This outcome is likely due to the major correlation between the "left" and "right" tasks, as both tasks aim to identify one out of the same ten digits. The first hidden layer (layer 0) exhibits a degree of similarity, likely arising from the overlap between the two digits. The following hidden layer (layer 1) shows the least similarity. The similarity then increases as moving deeper into the network. A possible explanation could be that early layers focus on the low-level image features at different image locations (left/right), while deeper layers focus on the high-level features for the classification of the identified digit.
\end{itemize}

The results show that the sub-networks are adjusted during the process, their separation can change at different layers and can depend on the similarity of the tasks.  

Therefore, the proposed method may offer some additional useful computational properties. In contrast to the compared baselines that require the full network for inference, our BU visual process is guided to operate only on a sparse sub-network, resulting in only a portion of the model that is used during inference. Consequently, after training, we can omit the unused parts of the model, resulting in a compact representation of the model, which is efficient both in terms of computation and memory. For example, we can drop approximately $80\%$ of the network when running the model in inference on Multi-MNIST. Furthermore, the compactness indicates the capacity of the model to accommodate a larger number of tasks within the same network.  

\subsubsection{Functional sparse sub-networks}

There has been a growing interest in the use of functional sparse sub-networks, following the Lottery-Ticket Hypothesis \cite{frankle2018lottery}. The hypothesis suggests that large dense networks contain smaller sub-networks that can be learned in isolation and match the performance of the full network on the learned task. This hypothesis has been supported by empirical evidence and was proven under certain conditions \cite{malach2020proving}. However, finding such sub-networks is challenging and is an active area of current research \cite{chen2021elastic, morcos2019one, ramanujan2020s, tanaka2020pruning, yu2022combinatorial}.

In this paper, we propose to extend this hypothesis suggesting that a sufficiently large network contains multiple overlapping sub-networks, each dedicated to a different task, resembling a sparse modular architecture. Our work suggests that these sub-networks can be naturally revealed by the same top-down mechanism used for propagating feedback signals in conventional networks (backpropagation). These findings, which are inspired by the observation of a unified top-down mechanism for both learning and guiding attention, highlight the potential benefits of the interactions between artificial intelligence and the human brain.

\end{document}